\documentclass[10pt,twocolumn,letterpaper]{article}

\usepackage[pagenumbers]{cvpr} 

\usepackage[accsupp]{axessibility}  

\newcommand{\methodname}{Blurry-Edges\xspace}

\usepackage{multirow}
\usepackage{makecell}
\usepackage{colortbl}
\usepackage{textcomp}

\newcommand{\citeS}[1]{\mbox{[\textcolor{cvprblue}{S}\citenum{#1}]}}

\definecolor{cvprblue}{rgb}{0.21,0.49,0.74}
\usepackage[pagebackref,breaklinks,colorlinks,allcolors=cvprblue]{hyperref}

\usepackage{physics}

\def\paperID{13831}
\def\confName{CVPR}
\def\confYear{2025}

\title{Blurry-Edges: Photon-Limited Depth Estimation from Defocused Boundaries}

\author{
Wei Xu, Charles James Wagner, Junjie Luo, and Qi Guo\\
Elmore Family School of Electrical and Computer Engineering, Purdue University\\
{\tt\small \{xu1639,wagne329,luo330,qiguo\}@purdue.edu}
}

\begin{document}

\twocolumn[{%
\renewcommand\twocolumn[1][]{#1}%
\maketitle
\begin{center}
    \centering
    \captionsetup{type=figure}
    \includegraphics[width=1.00\textwidth]{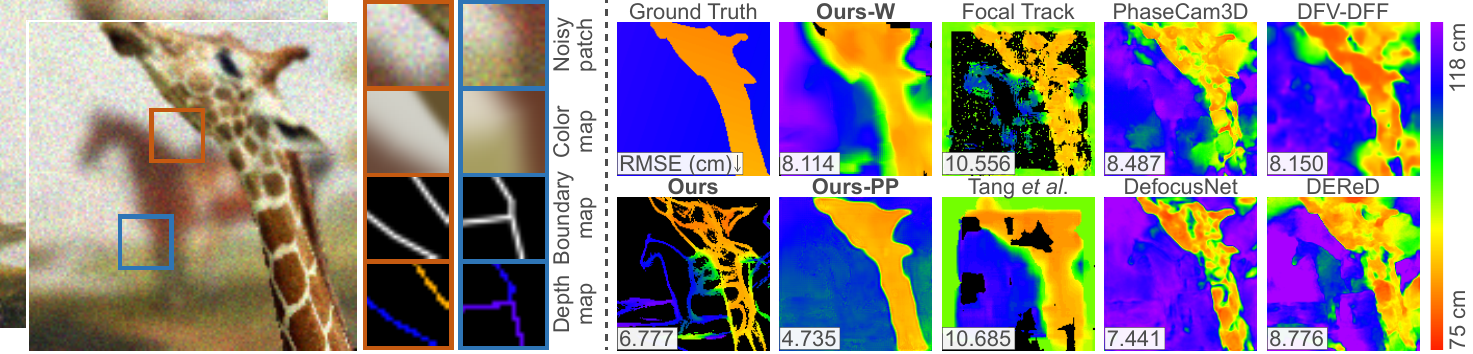}
    \captionof{figure}{Overview. (Left) \methodname~representation parametrically models an image patch's color, boundary positions, and boundary smoothness. Object depths can be analytically calculated from the smoothness of corresponding boundaries in a pair of differently defocused images. (Right) Compared to a variety of state-of-the-art depth from defocus algorithms~\cite{guo2017focal,tang2017depth,wu2019phasecam3d,maximov2020focus,yang2022deep,si2023fully}, our method generates sparse or dense depth maps with the lowest depth estimation errors from photon-limited, noisy images. }
    \label{fig:teaser}
\end{center}
}]

\begin{abstract}
Extracting depth information from photon-limited, defocused images is challenging because depth from defocus (DfD) relies on accurate estimation of defocus blur, which is fundamentally sensitive to image noise. We present a novel approach to robustly measure object depths from photon-limited images along the defocused boundaries. It is based on a new image patch representation, \methodname, that explicitly stores and visualizes a rich set of low-level patch information, including boundaries, color, and smoothness. We develop a deep neural network architecture that predicts the \methodname representation from a pair of differently defocused images, from which depth can be calculated using a closed-form DfD relation we derive. The experimental results on synthetic and real data show that our method achieves the highest depth estimation accuracy on photon-limited images compared to a broad range of state-of-the-art DfD methods. 

\end{abstract}
\section{Introduction}
\label{sec:intro}

Depth from defocus (DfD) generates physically accurate depth maps without additional, active illumination like time-of-flight or structured light~\cite{kumagai20217,koerner2021models,ou2023binocular,kim2022metasurface}, and has a monocular and compact form factor compared to stereo~\cite{lin2021depth,jacob2020depth}. These advantages make DfD suitable for spatially constrained artificial platforms, such as AR/VR, smartphones and watches, miniature robots, and drones. 

However, DfD relies on accurately estimating spatial derivatives in the captured images, a proxy of defocus level, as the depth cue, which is highly susceptible to the image noise~\cite{subbarao1997noise,anwar2021deblur,alexander2016focal}. To our knowledge, existing DfD solutions typically avoid this issue by assuming low noise levels in the input image (\cref{tab:related:noise}). Considering DfD's potential applications, which inevitably include dark environments, there is a pressing need for a DfD algorithm robust to photon-limited, noisy images. 

In light of this, we propose a method that robustly estimates object depth along the blurry boundaries from a pair of differently defocused noisy images. It leverages a novel patch structure representation named \textit{\methodname}. \methodname models an image patch as a stack of partially occluded wedges. As shown in \cref{fig:met:rep}, each wedge is parameterized by its vertex, color, and boundary blurriness. We develop a deep neural network to predict the optimal \methodname parameters that describe each patch and are consistent with neighboring patches' representation regarding boundary location, smoothness, and color. 

To perform depth estimation, our method utilizes a camera with a deformable lens to capture a pair of images of a static scene with varied focal lengths. The images share the same structure but have different smoothness at the boundaries due to the difference in defocus. By estimating the smoothness of the corresponding boundaries using \methodname, we can calculate the depth along the boundary from a closed-form DfD equation. 

We observe several critical advantages of the proposed DfD algorithm. First, it can be trained using naive, synthesized images with basic geometries and effectively estimate depths on real-world captured images without fine-tuning. 
Second, the \methodname representation is multifunctional. Besides the depth prediction, \methodname simultaneously generates a boundary map including edges of all smoothness and a noiseless color map. Last and most importantly, the proposed method demonstrates the unprecedented robustness of estimating depth from photon-limited images. The proposed method shows the highest accuracy in depth prediction using noisy, photon-limited input images compared to state-of-the-art DfD algorithms in both simulation and real-world experiments. 

The contribution of the paper includes:
\begin{enumerate}
    \item A parametrized representation, \methodname, that simultaneously models the color, boundary, and blurriness of a noisy image patch;
    \item A closed-form DfD equation that associates the smoothness of the corresponding boundaries in a pair of differently defocused images to the depth; 
    \item A deep neural network architecture that robustly estimates object depth along boundaries from a pair of defocused images, handling $4\times$ higher noise level (in standard deviations) than previous DfD algorithms (\cref{tab:related:noise});
    \item A comprehensive simulation and real-world analysis that proves the robustness of the proposed method's depth estimation under limited photons and its generalizability in training.
\end{enumerate}
All data and code of this work can be found in \linebreak\href{https://blurry-edges.qiguo.org/}{https://blurry-edges.qiguo.org/}.
\section{Related Work}
\label{sec:related}

Depth from defocus (DfD) was first proposed decades ago~\cite{pentland1987new}, and it has undergone rapid progress in the past decade thanks to the maturation and accessibility of various optical technologies, such as diffractive optical elements~\cite{ikoma2021depth}, deformable lenses~\cite{guo2017focal}, and metasurfaces~\cite{guo2019compact}. There are currently two complementary lines of research in DfD. The first utilizes analytical, non-learning-based solutions that estimate partially dense depth maps with minimal computational resources, and the second exploits learning-based models to produce high-quality, dense depth maps with a higher computational cost. 

\textbf{Analytical DfD algorithms} leverage the physical relationship between the image derivatives~\cite{huang2007evaluation,malik2008novel,nayar1994shape,watanabe1996real,thelen2008improvements} or local spatial frequency spectrum~\cite{lee2009reduced,huang2005robust,xie2006wavelet} and the depth. Theoretically, at least two images of the same scene captured with different focal planes are required to measure an object's depth without ambiguity~\cite{szeliski2022computer}. Recently, a special family of DfD algorithms, depth from differential defocus, demonstrates unprecedentedly low computational cost by leveraging simple, mathematical relationships between the differential change of image defocus and the object depth and is validated by real-world prototypes~\cite{alexander2016focal,guo2017focal,guo2019compact,luo2024depth}. Despite being computationally efficient, a fundamental drawback of these analytical DfD algorithms is the degeneracy, \ie, unreliable depth estimations at textureless regions of the images due to the lack of defocus cues~\cite{alexander2016focal,tang2017depth}. Fortunately, it is possible to predict where the degeneracies will happen given an image and the unreliable depth estimations in such areas can be removed from the final depth estimation~\cite{tang2017depth,guo2017focal,guo2019compact,luo2024depth}. 

\textbf{Learning-based DfD algorithms} utilize deep neural network architectures to learn the mapping from the defocused images to the depth values from data~\cite{maximov2020focus,yang2022deep,cs2018depthnet}. Compared to the analytical solutions, this class of methods achieves higher-quality, dense depth maps at higher computational costs. For example, a recent analytical DfD algorithm costs fewer than 1k floating point operations (FLOPs) per pixel~\cite{guo2019compact}, while a U-Net-based DfD algorithm uses 300k FLOPs per pixel~\cite{wu2019phasecam3d}. The learning-based DfD algorithms bypass the degeneracy issue by implicitly learning to fill depth values in textureless regions based on neighboring depth estimations. Thanks to recent advances in optical technologies, people have also incorporated the design of the blur kernel into the learning process so that the optical design and the DfD algorithm are optimized in an end-to-end fashion~\cite{wu2019phasecam3d,chang2019deep,tan20213d,ikoma2021depth}. The jointly-optimized systems typically demonstrate more accurate depth estimation than systems with pre-determined, fixed optics. 

\textbf{The sensitivity to image noise} is a fundamentally challenging problem in DfD. This is because the defocus information needs to be extracted from the spatial gradients of the images, which becomes increasingly sensitive to noise when the image defocus is significant~\cite{schechner2000depth}. As shown in \cref{tab:related:noise}, past DfD algorithms typically assume a relatively low noise level in their experiments. When necessary, these methods simply suppress the noise by averaging multiple frames~\cite{alexander2016focal} or binning pixels~\cite{guo2017focal}, and some use specially designed filters to locally attenuate the perturbation of the noise~\cite{subbarao1997noise, watanabe1998rational}. 

\begin{table}[htb]
\small
    \centering
    \begin{tabular}{@{}l@{\hskip 0.07in}c@{\hskip 0.07in}c@{\hskip 0.07in}c@{}}
        \toprule
        Method & Venue'Year & \makecell[c]{Noise SD \\ (LSB) $\uparrow$} & \makecell[c]{Illuminance \\ (lux) $\downarrow$} \\
        \midrule
        Focal Flow~\cite{alexander2016focal} & ECCV'2016 & 0.09--0.63 & 67,832--3,323,680 \\
        Tang \etal~\cite{tang2017depth} & CVPR'2017 & 1.50--3.75 & 1,916--11,967 \\
        Focal Track~\cite{guo2017focal} & ICCV'2017 & 0.30--2.00 & 6,732--299,133 \\
        PhaseCam3D~\cite{wu2019phasecam3d} & ICCP'2019 & 2.55 & 4,142 \\
        Guo \etal~\cite{guo2019compact} & PNAS'2019 & 0.70 & 54,944 \\
        DefocusNet~\cite{maximov2020focus} & CVPR'2020 & 1.00--4.00 & 1,684--26,923 \\
        DEReD~\cite{si2023fully} & CVPR'2023 & 1.00--4.00 & 1,684--26,923 \\
        \textbf{Ours} & - & 18.21--19.22 & 74--83 \\
        \bottomrule
    \end{tabular}
    \caption{Image noise of previous DfD work. We convert the noise levels reported by each paper into the standard deviation (SD) in the unit of least significant bit (LSB) for 8-bit images. Images used in this work have at least $4 \times$ more significant noise. We also convert the Noise SD to the illuminance under common camera parameters, with calculation details in the supplementary. Images used in this work roughly correspond to photos taken under the twilight or a very dark day~\cite{enwiki:1243679454}.}
    \label{tab:related:noise}
\end{table}

In recent years, a series of works have utilized a novel patch representation, field-of-junction (FoJ), to regularize boundary detection from images~\cite{verbin2021field,polansky2024boundary,xu2024ct}. FoJ demonstrates extraordinary robustness in detecting boundary structures from images at an extremely low signal-to-noise ratio, as restricting the variety of local patch structures can effectively attenuate the impact of noise in image restoration~\cite{ofir2019detection}. However, FoJ does not model boundary smoothness, and the boundary structures it can represent are limited to lines, edges, and junctions. If a more general patch representation incorporating boundary smoothness and more sophisticated boundary structures can be developed, it could be utilized to detect the defocus along boundaries robustly in the presence of significant noise.
\section{Methods}
\label{sec:met}

\subsection{Depth from Defocus}
\label{sec:met:defocus}

Consider a wide-aperture lens imaging a front parallel target.
Under paraxial approximation, the captured image on the photosensor is mathematically the convolution of the point spread function (PSF) $k(\boldsymbol{x})$ and the pinhole image $Q(\boldsymbol{x})$:
\begin{equation}
    I \left( \boldsymbol{x} \right) = Q \left( \boldsymbol{x} \right) \ast k \left( \boldsymbol{x}, \sigma \left( z \right) \right),
    \label{eq:met:cam}
\end{equation}
where $\boldsymbol{x}$ is the 2D position on the photosensor. Assuming the PSF has a Gaussian intensity profile and the defocus process follows the thin lens law, the PSF $k(\boldsymbol{x})$ can be mathematically expressed as:
\begin{equation}
    k \left( \boldsymbol{x}, \sigma \left( z \right) \right) = \frac{1}{2 \pi \left( \sigma \left( z \right) \right)^2} \exp \left( - \frac{\left\|\boldsymbol{x}\right\|^2}{2 \left( \sigma \left( z \right) \right)^2} \right),
\end{equation}
where the defocus level $\sigma \left( z \right)$ is determined by the target's depth $z$ and constant parameters of the optical system~\cite{guo2017focal}:
\begin{equation}
    \sigma \left( z \right) = \Sigma \left[ \left( \frac{1}{z} - \rho \right) s + 1 \right],
    \label{eq:met:thin_lens}
\end{equation}
where $\Sigma$ represents the standard deviation of the Gaussian aperture function, $\rho$ is the dioptric power of the lens, and $s$ is the separation between the photosensor and the lens.

Now we consider the textures in the pinhole image $Q(\boldsymbol{x})$. To approximate the textures of different sharpness, we model each small patch $P$ of the pinhole image $Q(\boldsymbol{x})$ as the convolution of a Gaussian kernel $k(\boldsymbol{x}; \xi)$ with standard deviation $\xi$ and a piecewise 2D step function $\bar{Q}(\boldsymbol{x})$:
\begin{equation}
    Q \left( \boldsymbol{x} \right) = \bar{Q} \left( \boldsymbol{x} \right) \ast k \left( \boldsymbol{x}, \xi \right), \boldsymbol{x} \in P.
    \label{eq:met:texsmoothness}
\end{equation}
For sharp textures, the Gaussian kernel has a relatively small standard deviation $\xi$, and vice versa. Combining \cref{eq:met:texsmoothness} with \cref{eq:met:cam}, the captured image $I(\boldsymbol{x})$ can be represented as:
\begin{equation}
    I \left(\boldsymbol{x}\right) = \bar{Q}(\boldsymbol{x}) \ast k\left(\boldsymbol{x}, \sqrt{\sigma(z)^2 + \xi^2}\right),
    \label{eq:met:imgrender}
\end{equation}
where the term $\sqrt{\sigma(z)^2 + \xi^2}$ indicates the smoothness value of the boundaries in the patch $P$. 

Consider a deformable lens that can dynamically vary its optical power, with a visualization provided in the supplementary. The system can sequentially capture two images of a static scene, $I_+$ and $I_-$, with different optical powers, $\rho_+$ and $\rho_-$. By estimating the smoothness value of a corresponding boundary in a patch $P$, $\eta_+$ and $\eta_-$, we have the mathematical relationships:
\begin{equation}
    \sqrt{\eta_\pm^2 - \xi^2} = \Sigma \left[\left(\frac{1}{z} - \rho_\pm\right)s + 1\right].
\end{equation}
By combining both equations to cancel out $\xi$, we obtain the following equation to calculate the depth of the boundary given a pair of estimated smoothness $\eta_+$ and $\eta_-$:
\begin{equation}
    z \hspace{-0.01in} \left( \hspace{-0.007in} \eta_+, \hspace{-0.01in} \eta_- \hspace{-0.007in} \right) \hspace{-0.021in} = \hspace{-0.021in} \frac{2 \Sigma^2 s^2 \hspace{-0.023in} \left( \hspace{-0.007in} \rho_- \hspace{-0.023in} - \hspace{-0.023in} \rho_+ \hspace{-0.007in} \right)}{\eta_+^2 \hspace{-0.023in} - \hspace{-0.023in} \eta_-^2 - \hspace{-0.023in} \Sigma^2 s \hspace{-0.023in} \left( \hspace{-0.007in} \rho_+ \hspace{-0.023in} - \hspace{-0.023in} \rho_- \hspace{-0.007in} \right) \hspace{-0.023in} \left( \hspace{-0.007in} s \rho_+ \hspace{-0.023in} + \hspace{-0.023in} s \rho_- \hspace{-0.023in} - \hspace{-0.023in} 2 \hspace{-0.007in} \right)}\hspace{-0.016in}.
    \label{eq:met:depthsol}
\end{equation}

\subsection{\methodname Representation}
\label{sec:met:rep}

\begin{figure}
    \centering
    \includegraphics[width=0.81\linewidth]{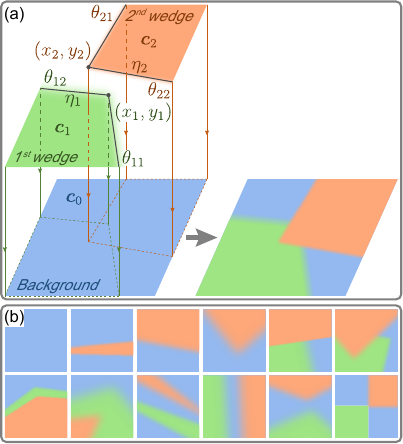}
    \caption{\methodname representation with the number of wedges $l=2$. (a) The $i$th wedge is parameterized by the vertex position $(x_i, y_i)$, the starting and ending angle $(\theta_{i1}, \theta_{i2})$, the color $\boldsymbol{c}_i$, and the boundary smoothness $\eta_i$. The rendering of the patch is through the alpha compositing of the wedges. (b) \methodname can represent a variety of boundary structures. In particular, it can represent structures with various boundary smoothness.}
    \label{fig:met:rep}
\end{figure}

\methodname represents an image patch as the alpha compositing of $l$ vertically-stacked, constant-color wedges with smooth boundaries. As illustrated in \cref{fig:met:rep}a, each patch is modeled by a set of parameters, 
\begin{equation}
    \boldsymbol{\Psi} = \left( \{\boldsymbol{p}_i, \boldsymbol{\theta}_i, \boldsymbol{c}_i, \eta_i, i = 1,2,\cdots,l\}, \boldsymbol{c}_0\right).
\end{equation}
The tuple $(\boldsymbol{p}_i, \boldsymbol{\theta}_i, \boldsymbol{c}_i, \eta_i)$ parameterize the $i$th wedge in the patch, with $\boldsymbol{p}_i = (x_i, y_i)$ representing the vertex, $\boldsymbol{\theta}_i = (\theta_{i1}, \theta_{i2})$ denoting the starting and ending angle, $\boldsymbol{c}_i$ indicating the RGB color, and $\eta_i$ recording the smoothness of the boundary. The wedge with a large index is in the front. The vector $\boldsymbol{c}_0$ represents the RGB color of the background. As shown in \cref{fig:met:rep}b, this representation can model various boundary structures and smoothness.

\begin{figure}[htb]
    \centering
    \includegraphics[width=0.60\linewidth]{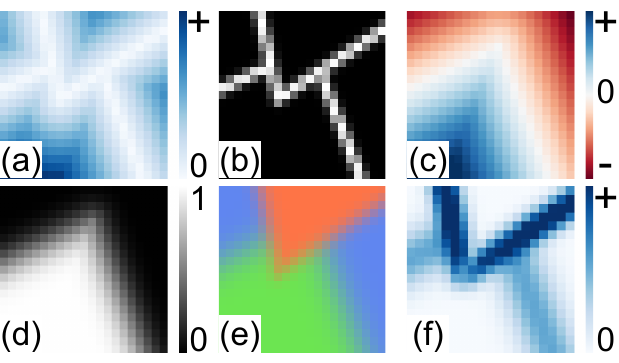}
    \caption{Visualizations from a sample \methodname representation. (a) The unsigned distance map to the nearest unoccluded boundary, $u \left( \boldsymbol{x}; \boldsymbol{\Psi} \right)$. (b) The corresponding boundary center map, $b \left( \boldsymbol{x}; \boldsymbol{\Psi}, \delta \right)$. (c) The signed distance map of the bottom wedge, $d_1 \left( \boldsymbol{x}; \boldsymbol{\Psi} \right)$. (d) The $\alpha$-map of the bottom wedge, $\alpha_1 \left( \boldsymbol{x}; \boldsymbol{\Psi} \right)$. (e) The color map of the patch, $c \left( \boldsymbol{x}; \boldsymbol{\Psi} \right)$. (f) The magnitude of color derivative map of the patch, $c^{\prime} \left( \boldsymbol{x}; \boldsymbol{\Psi} \right)$.}
    \label{fig:met:dist}
\end{figure}

Given a \methodname representation of a patch $\boldsymbol{\Psi}$, several types of auxiliary visualizations can be generated. First, the boundary center map $b \left( \boldsymbol{x}; \boldsymbol{\Psi}, \delta \right)$ highlights the center of each unoccluded boundary in the patch (\cref{fig:met:dist}b.) It is computed via:
\begin{equation}
    b \left( \boldsymbol{x}; \boldsymbol{\Psi}, \delta \right) = \exp \left[ - \frac{\left( u \left( \boldsymbol{x}; \boldsymbol{\Psi} \right) \right)^2}{\delta^2} \right],
    \label{eq:met:bndrypatch}
\end{equation}
where $\delta$ is a hyperparameter that controls the stroke of the visualized boundaries and $u \left( \boldsymbol{x}; \boldsymbol{\Psi} \right)$ is an unsigned distance map to the nearest unoccluded boundary center for each pixel. The exact calculation of the distance map can be found in the supplementary. \Cref{fig:met:dist}a shows the distance map to generate \cref{fig:met:dist}b.

Second, the color map $c\left( \boldsymbol{x}; \boldsymbol{\Psi} \right)$ is the rendering of the stacked, colored wedges according to their boundary smoothness and occlusion. It can be computed via alpha compositing:
\begin{equation}
    c \left( \boldsymbol{x}; \boldsymbol{\Psi} \right) = \sum_{i=0}^l \boldsymbol{c}_i \alpha_{l\rightarrow i} \left( \boldsymbol{x}; \boldsymbol{\Psi} \right),
    \label{eq:met:colorpatch}
\end{equation}
where $\alpha_{l\rightarrow i} \left( \boldsymbol{x}; \boldsymbol{\Psi} \right)$ is the collective $\alpha$-map from the $l$th to the $i$th wedge:
\begin{equation}
    \alpha_{l\rightarrow i} \left( \boldsymbol{x}; \boldsymbol{\Psi} \right) = \alpha_i \left( \boldsymbol{x}; \boldsymbol{\Psi} \right) \prod_{j=i+1}^l \left( 1 - \alpha_j \left( \boldsymbol{x}; \boldsymbol{\Psi} \right) \right).
    \label{eq:met:weight}
\end{equation}
The term $\alpha_i \left( \boldsymbol{x}; \boldsymbol{\Psi} \right)$ is the $\alpha$-map of the $i$th wedge:
\begin{equation}
    \alpha_i \left( \boldsymbol{x}; \boldsymbol{\Psi} \right) = \frac{1}{2} \left[ 1 + \erf \left( \frac{d_i \left( \boldsymbol{x}; \boldsymbol{\Psi} \right)}{ \sqrt{2} \eta_i} \right) \right],
\end{equation}
where $\erf(\cdot)$ indicates the Gausian error function, $d_i \left( \boldsymbol{x}; \boldsymbol{\Psi} \right)$ denotes the signed distance map of the $i$th wedge (\cref{fig:met:dist}c), and $\eta_i$ is the boundary smoothness of the wedge. \Cref{fig:met:dist}e shows a sample color map that corresponds to the boundary center map in \cref{fig:met:dist}b.

Besides the boundary center map and the color map, \methodname also enables a color derivative map $c'\left( \boldsymbol{x}; \boldsymbol{\Psi} \right)$ that highlights the boundary smoothness. We compute the color derivative map as the color map's response to the Sobel operator~\cite{sobel2022sobel}:
\begin{equation}
    c^{\prime} \left( \boldsymbol{x}; \boldsymbol{\Psi} \right) \hspace{-0.034in} = \hspace{-0.034in} \sqrt{\left( c \left( \boldsymbol{x}; \boldsymbol{\Psi} \right) \ast G_x \right)^2 \hspace{-0.034in} + \hspace{-0.034in} \left( c \left( \boldsymbol{x}; \boldsymbol{\Psi} \right) \ast G_y \right)^2},
    \label{eq:met:derivative}
\end{equation}
where $G_x$ and $G_y$ are the Sobel kernels in x and y directions. A sample color derivative map is visualized in \cref{fig:met:dist}f.

\subsection{Depth estimation}
\label{sec:met:net}

\begin{figure*}[htb]
    \centering
    \includegraphics[width=0.81\linewidth]{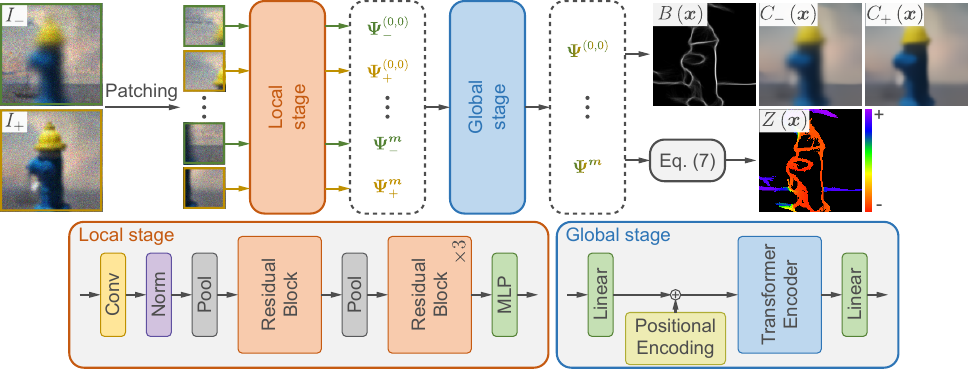}
    \caption{Framework of the proposed model. There are two stages. The local stage consists of residual blocks and predicts the \methodname representation for each patch locally. The global stage consists of a Transformer Encoder and refines the \methodname representation for all patches globally. Finally, the framework combines all the per-patch representations and outputs the global boundary map, color map, and depth map.}
    \label{fig:met:framework}
\end{figure*}

\Cref{fig:met:framework} shows our DfD algorithm based on the \methodname representation. The input is a pair of differently defocused images of a static scene, $I_+, I_- \in \mathbb{R}^{H \times W \times k}$. For simplicity of notation, we use $I_\pm$ to represent the pair of images throughout the paper. The model first estimates the \methodname representation of the images in two stages and then generates the depth map from it. 

First, our method divides the images into uniform-size, overlapping patches and independently predicts the \methodname representation of each patch using a convolutional neural network (CNN) based architecture. Given a patch, $P\in \mathbb{R}^{h \times w \times k}$, the CNN predicts a part of its \methodname representation, including the vertex locations $\left\{\boldsymbol{p}_i\right\}$, the angles $\left\{\boldsymbol{\theta}_i\right\}$, and the boundary smoothness $\left\{\eta_i\right\}$. Then, it computes the color information $\left\{\boldsymbol{c}_i\right\}$ using these predicted parameters and the patch $P$ via ridge regression:
\begin{equation}
     \begin{bmatrix}
         \boldsymbol{c}_0 \\
         \vdots \\
         \boldsymbol{c}_l
     \end{bmatrix} \hspace{-0.03in} = \hspace{-0.03in}
     \left(A^{\top} \hspace{-0.02in} A \hspace{-0.02in} + \hspace{-0.02in} \lambda \mathbb{I}_{(l+1)\times (l+1)}\right)^{-1} \hspace{-0.05in}
     \begin{bmatrix}
         \alpha_{l\rightarrow 0}(\boldsymbol{x}) \hspace{-0.03in} \cdot \hspace{-0.03in} P(\boldsymbol{x}) \\
         \vdots \\
         \alpha_{l\rightarrow l}(\boldsymbol{x}) \hspace{-0.03in} \cdot \hspace{-0.03in} P(\boldsymbol{x})
     \end{bmatrix}\hspace{-0.03in},
     \label{eq:met:ridge}
\end{equation}
where $\alpha_{l\rightarrow 0}\left(\boldsymbol{x}\right)$ is the collective $\alpha$-map described in \cref{eq:met:weight} and $\alpha_{l\rightarrow i}\left(\boldsymbol{x}\right) \cdot P\left(\boldsymbol{x}\right)$ indicate the channel-wise dot product between the collective $\alpha$-map and the patch, and the matrix $A = \begin{bmatrix}
    \alpha_{l\rightarrow 0}(\boldsymbol{x}) \cdots \alpha_{l\rightarrow l}(\boldsymbol{x})
\end{bmatrix}$. 

The first stage estimates \methodname representation of each patch purely based on the local information. Thus, we refer to it as the \textit{local} stage. For notation purposes, we denote a patch cropped from one of the two images, $I_{\pm}$, as $P_{\pm}^{\boldsymbol{m}}$. The subscript $\pm$ represents the image from which it is cropped, and the superscript $\boldsymbol{m}=(m,n)$ indicates the center position of the patch from the original image. Its \methodname representation predicted by the local stage is denoted as $\boldsymbol{\Psi}_{\pm}^{\boldsymbol{m}}$, as shown in \cref{fig:met:framework}.

In the second or the \textit{global} stage, the model leverages a Transformer Encoder to take in all \methodname representations, $\left\{\boldsymbol{\Psi}_{\pm}^{\boldsymbol{m}}, \forall \boldsymbol{m}\right\}$ and refine them according to several consistency constraints. First, for each pair of patches corresponding to the center position $\boldsymbol{m}$ in the pair of images, the global stage outputs a regularized \methodname representation, $\boldsymbol{\Psi}^{\boldsymbol{m}} = \left(\boldsymbol{\Omega}^{\boldsymbol{m}}, \boldsymbol{\eta}^{\boldsymbol{m}}_+, \boldsymbol{\eta}^{\boldsymbol{m}}_- \right)$:
\begin{equation}
    \begin{aligned}
    \boldsymbol{\Omega}^{\boldsymbol{m}} =& \left\{\boldsymbol{p}_i, \boldsymbol{\theta}_i, \boldsymbol{c}_i, {\boldsymbol{c}}_0, i = 1,\cdots,l \right\}, \\
    \boldsymbol{\eta}^{\boldsymbol{m}}_+ =& \left\{\eta_{i,+}, i = 1,\cdots,l \right\}, \\
    \boldsymbol{\eta}^{\boldsymbol{m}}_- =& \left\{\eta_{i,-}, i = 1,\cdots,l \right\}. 
    \end{aligned}
\end{equation}
This regularized \methodname representation enforces the \textit{defocus consistency}, \ie, the pair of patches share the same wedge positions and colors, $\boldsymbol{\Omega}^{\boldsymbol{m}}$, but different boundary smoothness as specified by $\boldsymbol{\eta}^{\boldsymbol{m}}_+$ and $\boldsymbol{\eta}^{\boldsymbol{m}}_-$. Then, using the DfD equation~(\cref{eq:met:depthsol}), the depth value of each wedge can be solved from the two corresponding smoothness values $\eta_{i,+}, \eta_{i,-}$:
\begin{equation}
    z_i^{\boldsymbol{m}} = z \left( \eta_{i,+}, \eta_{i,-} \right).
    \label{eq:met:depth-smooth}
\end{equation}

The Transformer Encoder is trained to also promote consistency among neighboring patches in terms of boundary center maps, color maps, and color derivative maps. The loss functions to be used to promote these consistencies will be discussed in \cref{sec:met:train}. More details of the network architecture can be found in the supplementary.

\begin{figure}[htb]
    \centering
    \includegraphics[width=0.77\linewidth]{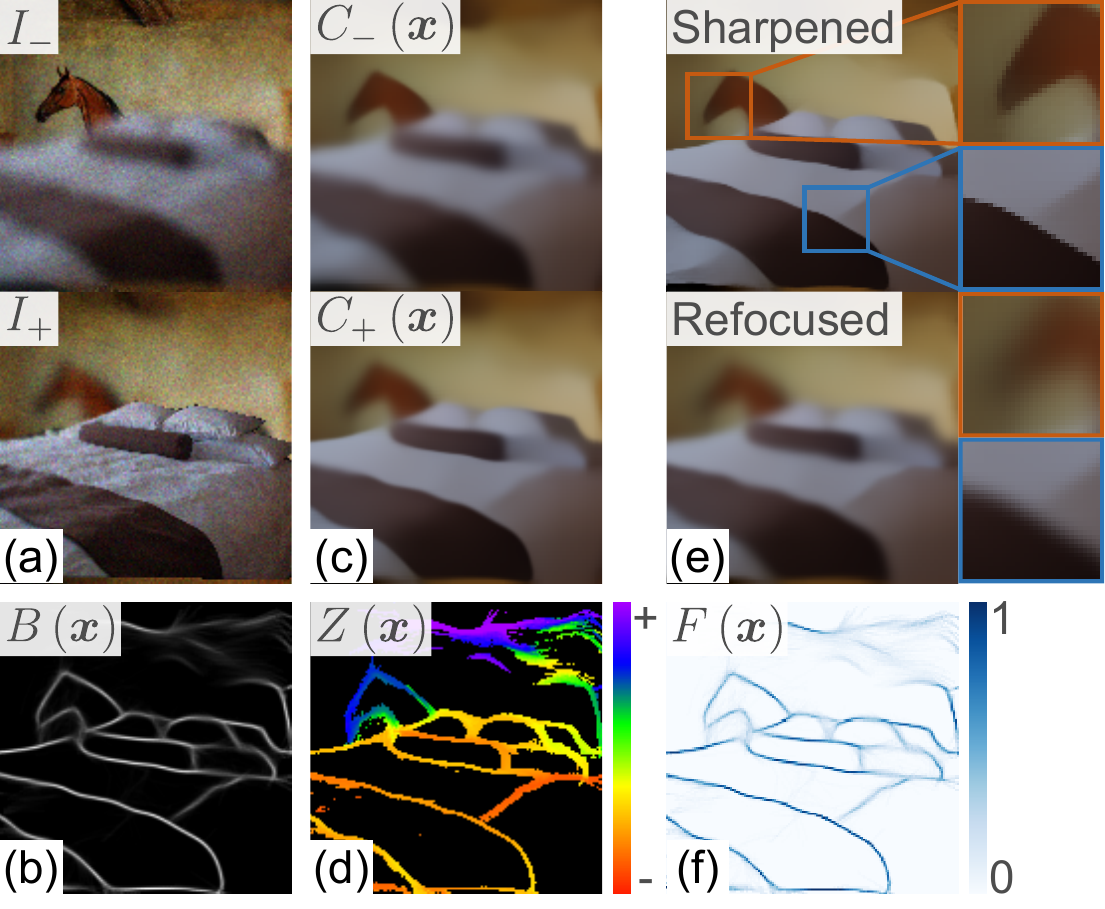}
    \caption{Examples of inputs and global outputs. (a) Noisy input image pair $I_\pm$ with different optical power $\rho_\pm$. (b) Global boundary center map $B \left( \boldsymbol{x} \right)$. (c) Global color map $C \left( \boldsymbol{x} \right)$. (d) Global sparse depth map $Z \left(\boldsymbol{x} \right)$. (e) Sharpened and refocused color maps. (f) Global confidence map $F \left( \boldsymbol{x} \right)$.}
    \label{fig:met:output}
\end{figure}

Finally, the model calculates a global boundary center map, a global color map, and a global depth image by aggregating all patchwise \methodname representations. The global boundary center map $B \left( \boldsymbol{x} \right)$ is computed by averaging all per-patch boundary maps:
\begin{equation}
    B \left( \boldsymbol{x} \right) = \frac{1}{\left|P^{\boldsymbol{m}}_{\pm} \ni \boldsymbol{x}\right|} \sum_{P^{\boldsymbol{m}}_{\pm} \ni \boldsymbol{x}} 
    b \left( \boldsymbol{x} - \boldsymbol{m}; \boldsymbol{\Omega}^{\boldsymbol{m}}, \delta \right),
\end{equation}
where $P^{\boldsymbol{m}}_{\pm} \ni \boldsymbol{x}$ indicates all patches centered at $\boldsymbol{m}$ that contain pixel $\boldsymbol{x}$ and $\left|P^{\boldsymbol{m}}_{\pm} \ni \boldsymbol{x}\right|$ denotes the number of such patches. The global color map $C \left( \boldsymbol{x} \right)$ is computed similarly by averaging the local color maps, but it can be augmented with different smoothness values for each wedge:
\begin{equation}
    C \left( \boldsymbol{x} \right) = \frac{1}{\left|P^{\boldsymbol{m}}_{\pm} \ni \boldsymbol{x}\right|} \sum_{P^{\boldsymbol{m}}_{\pm} \ni \boldsymbol{x}} 
    c \left( \boldsymbol{x} - \boldsymbol{m}; \left\{\boldsymbol{\Omega}^{\boldsymbol{m}}, \boldsymbol{\eta}^{\boldsymbol{m}}\right\}\right).
\end{equation}
The parameter $\boldsymbol{\eta}^{\boldsymbol{m}}$ denotes the smoothness values for all wedges in the patch. When setting the smoothness value $\boldsymbol{\eta}^{\boldsymbol{m}} = \boldsymbol{\eta}^{\boldsymbol{m}}_{\pm}$, the generated color maps correspond to the input image pairs $I_\pm$, which are $C_\pm \left( \boldsymbol{x} \right)$. Furthermore, the model can generate a refocused or sharpened color map by setting $\boldsymbol{\eta}^{\boldsymbol{m}}$ to different values. Examples are shown in \cref{fig:met:output}e. From the global color map $C(\boldsymbol{x})$, we can also calculate the global color-derivative map $C'(\boldsymbol{x})$ by performing the Sobel filtering as in \cref{eq:met:derivative}. 

The global sparse depth map $Z \left(\boldsymbol{x} \right)$ visualizes the depth values along the boundary centers:
\begin{equation}
    Z \hspace{-0.025in} \left( \boldsymbol{x} \right) \hspace{-0.037in} = \hspace{-0.037in} \frac{\sum_{P^{\boldsymbol{m}}_{\pm} \ni \boldsymbol{x}} \hspace{-0.025in} \sum_{i=1}^l \hspace{-0.02in}
    H \hspace{-0.025in} \left( b_i \hspace{-0.025in} \left( \boldsymbol{x} \hspace{-0.025in} - \hspace{-0.025in} \boldsymbol{m}; \hspace{-0.01in} \boldsymbol{\Omega}^{\boldsymbol{m}}, \hspace{-0.01in} \delta \right) \hspace{-0.025in} - \hspace{-0.025in} \tau \right) \hspace{-0.03in} \cdot \hspace{-0.03in} z_i^{\boldsymbol{m}}}{\sum_{P^{\boldsymbol{m}}_{\pm} \ni \boldsymbol{x}} \hspace{-0.02in} 
    H \hspace{-0.025in} \left( b \hspace{-0.025in} \left( \boldsymbol{x} \hspace{-0.025in} - \hspace{-0.025in} \boldsymbol{m}; \hspace{-0.01in} \boldsymbol{\Omega}^{\boldsymbol{m}}, \hspace{-0.01in} \delta \right) \hspace{-0.025in} - \hspace{-0.025in} \tau \right)}\hspace{-0.022in},
\end{equation}
where $H\left(\cdot\right)$ is the Heaviside step function, $b_i \left( \boldsymbol{x}; \boldsymbol{\Omega}^{\boldsymbol{m}}, \delta \right) = b \left( \boldsymbol{x}; \boldsymbol{\Omega}^{\boldsymbol{m}}, \delta \right) \cdot M_i \left( \boldsymbol{x} \right)$ is the unoccluded boundary center of the $i$th wedge ($M_i \left( \boldsymbol{x} \right)$ is the mask for the unoccluded $i$th wedge, whose calculation is in the supplementary), $\tau$ is a hyperparameter to control the coverage of the depth value, $z_i^{\boldsymbol{m}}$ is the estimated depth value of the $i$th wedge in patch $P^{\boldsymbol{m}}_{\pm}$ according to \cref{eq:met:depth-smooth}. The model also outputs a global confidence map that predicts and filters unreliable boundary and depth estimations. It is calculated via:
\begin{equation}
    F \left( \boldsymbol{x} \right) \hspace{-0.021in} = \hspace{-0.021in} \frac{1}{\left|P^{\boldsymbol{m}}_{\pm} \hspace{-0.015in} \ni \hspace{-0.015in} \boldsymbol{x}\right|} \hspace{-0.03in} \sum_{P^{\boldsymbol{m}}_{\pm} \ni \boldsymbol{x}} \hspace{-0.03in}
    H \left( b \left( \boldsymbol{x} \hspace{-0.01in} - \hspace{-0.01in} \boldsymbol{m}; \boldsymbol{\Omega}^{\boldsymbol{m}}, \delta \right) \hspace{-0.01in} - \hspace{-0.01in} \tau \right)\hspace{-0.01in}.
\end{equation}
\Cref{fig:met:output} visualizes the global maps generated from our DfD algorithm of a sample synthesized scene.

\subsection{Training}
\label{sec:met:train}

We design a modular learning scheme that trains the local and global stages of our model independently. First, we optimize the parameters of the CNN in the local stage using the following loss function:
\begin{equation}
    \mathcal{L}_{\text{local}} = \sum_{i=1}^3 \beta_i \mathbb{E}_{\boldsymbol{m}}\left(l_i\right),
    \label{eq:met:local-obj}
\end{equation}
where $\mathbb{E}_{\boldsymbol{m}}$ denotes the expectation over all patches in an image. The loss function consists of three terms, $l_i$, that comprehensively penalize the color error, smoothness error, and boundary localization error. After the local stage converges, we train the Transformer Encoder in the global stage with a fixed local stage using a comprehensive loss function that consists of seven terms:
\begin{equation}
    \mathcal{L}_{\text{global}} = \sum_{i=1}^7 \gamma_i \mathbb{E}_{I_\pm, \boldsymbol{m}}\left(g_i\right),
    \label{eq:met:global-obj}
\end{equation}
where $\mathbb{E}_{I_\pm, \boldsymbol{m}}$ denotes the expectation over all image pairs $I_\pm$ in the training set and all corresponding patches of each image pair. The seven loss terms, $g_i$, comprehensively penalize the prediction error and inconsistency among neighboring patches regarding colors, boundary locations, boundary smoothness, and depth. The exact derivation is shown in the supplementary. During the training of the two stages, we observe that dynamically changing the coefficients $\beta_i$, $\gamma_i$ helps with the convergence, which is also discussed in the supplementary. We will describe other details of the training configurations in \cref{sec:experiment:config}.
\section{Experimental Results}
\label{sec:experiment}

\subsection{Training Configurations}
\label{sec:experiment:config}

We fix the number of wedges $l=2$ throughout the experiments, providing the optimal balance between accuracy and computational complexity from our experience. The framework is implemented in PyTorch~\cite{paszke2017automatic}. We use the AdamW optimizer~\cite{loshchilov2017decoupled} and the ReduceLROnPlateau scheduler for training both stages. The initial learning rates are $6 \times 10^{-5}$ and $1 \times 10^{-4}$ for local and global stages. The two stages are trained with batch sizes of 64 and 8 for 1000 and 350 epochs, respectively. We provide a more detailed description of the training parameters in the supplementary. The training and testing are performed on an NVIDIA GeForce RTX A5000 graphics card with 24 GB of memory. 

\subsection{Datasets}
\label{sec:experiment:dataset}

The training set we generate consists of images with only basic geometries, \ie, rectangles, circles, and triangles. Each object has a constant, random depth value ranging from 0.75 m to 1.18 m. We apply the Poisson-Gaussian noise to the synthesized images~\cite{ding2016modeling}:
\begin{equation}
    I \left( \boldsymbol{\mathbf{x}} \right) = \mathrm{Poisson} \left( \alpha I^{\ast} \left( \boldsymbol{\mathbf{x}} \right) \right) + \mathrm{Gaussian} \left( 0, \sigma^2 \right),
\end{equation}
where $I \left( \boldsymbol{\mathbf{x}} \right)$ and $I^{\ast} \left( \boldsymbol{\mathbf{x}} \right) \in \left[0, 1\right]$ are the noisy and normalized clean images, $\alpha \in \left[180, 200\right]$ is the photon level that controls the maximum photon capacity for each pixel, and $\sigma = 2$ is the standard deviation of the Gaussian read noise. We synthesize two images for each scene with optical powers $\rho_- = 10.0~\mathrm{m}^{-1}$ and $\rho_+ = 10.2~\mathrm{m}^{-1}$. 

\begin{figure}[htb]
    \centering
    \includegraphics[width=1.00\linewidth]{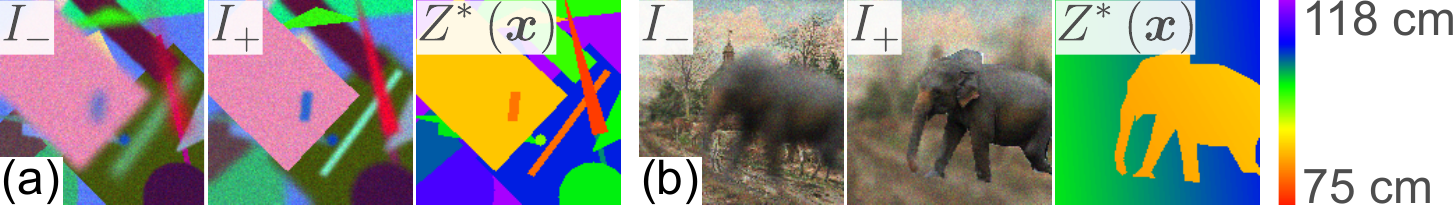}
    \caption{Sample images of the synthetic dataset. $Z^{\ast} \left( \boldsymbol{x} \right)$ indicates the ground truth depth map. (a) The training and validation set consists of basic geometries as objects. (b) The testing set contains objects with realistic textures and boundaries.}
    \label{fig:exp:data}
\end{figure}

The training and validation sets contain 8,000 and 2,000 randomly generated scenes, respectively. Sample images and the corresponding depth map are shown in \cref{fig:exp:data}a. For the local stage, we randomly cropped 16,000 and 4,000 patch samples from the training and the validation sets with significant boundaries for training and validation. We use the full images from these sets for the global stage. 
For the testing set, we avoid commonly used RGBD datasets, such as NYUDv2~\cite{silberman2012indoor}, as realistically rendering a defocused depth boundary requires the occluded background information that these datasets do not provide. 
Instead, we independently select foreground and background images from two image datasets. The background is randomly selected from the Painting dataset~\cite{crowley2015search}, and the foreground uses images from the MS-COCO dataset~\cite{lin2014microsoft}. We directly utilize the segmentation mask in the MS-COCO to create a foreground object with sophisticated textures and boundary shapes. Both the foreground and the background can have continuously changing depth values. We also follow the rendering framework of Guo \etal~\cite{guo2019compact}, which uses interpolated PSFs to create a smoothly changing defocus and alpha compositing for realistic depth boundaries. We render 200 scenes for the testing set. Sample images and the corresponding depth maps are shown in \cref{fig:exp:data}b. The images in our dataset have a resolution of $147 \times 147$ pixels.

\subsection{Patch size}
\label{sec:experiment:ablation}

\begin{figure}[htb]
    \centering
    \includegraphics[width=0.91\linewidth]{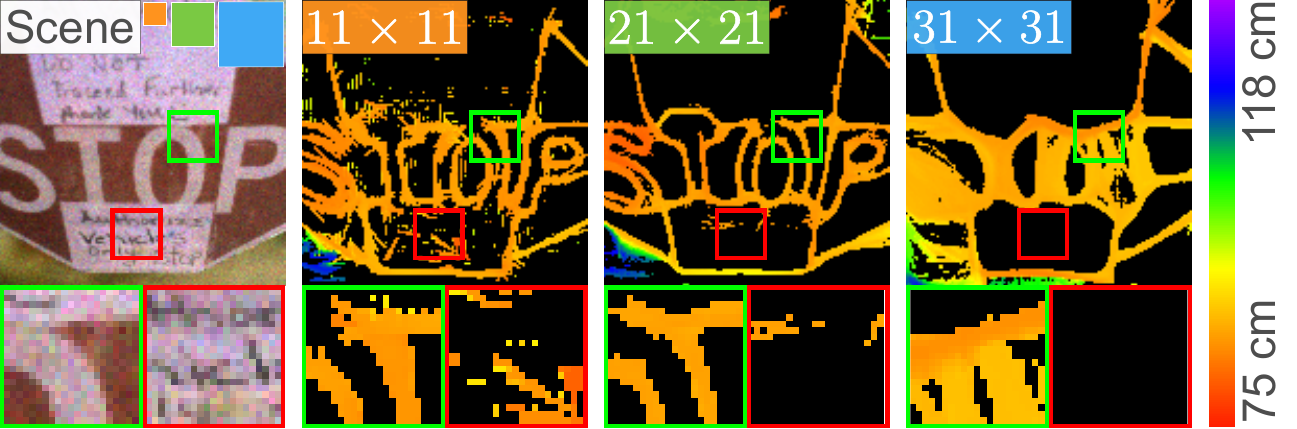}
    \caption{Depth map generated with different patch sizes, with patch sizes indicated by colored squares, \textcolor{RedOrange}{$11 \times 11$}, \textcolor{Green}{$21 \times 21$}, and \textcolor{Cerulean}{$31 \times 31$}, respectively. Our method estimates depth values along the boundaries. The patch size $21 \times 21$ results in the most accurate depth estimation with fine structures preserved. {$11 \times 11$ can detect tiny structures but requires a smaller stride for consistency}.}
    \label{fig:exp:ablation}
\end{figure}

The patch size is a critical hyperparameter of \methodname that impacts our algorithm's accuracy. To analyze the effect of the patch size and determine the optimal value, we train our model with three patch sizes, $11 \times 11$, $21 \times 21$, and $31 \times 31$, and quantitatively and qualitatively compare the depth prediction accuracy. As shown in \cref{tab:exp:ablation}, the patch size $21 \times 21$ achieves the highest accuracy across key metrics ($\delta1$, RMSE, and AbsRel) on the testing set. This can be intuitively explained from \cref{fig:exp:ablation}, where the patch size $21 \times 21$ strikes a balance between containing sufficient pixels for accurate depth estimation and retaining detailed structures in the image. Although a smaller patch size enables depth estimation along the tiny textures, it also requires a smaller stride for consistency, which increases memory usage. Therefore, we select the patch size of $21 \times 21$ and the stride of 2 after balancing the accuracy and computational efficiency throughout the experiment.

\begin{table}[htb]
    \centering
    \begin{tabular}{@{}l@{\hskip 0.03in}|@{\hskip 0.03in}c@{\hskip 0.05in}c@{\hskip 0.05in}c@{\hskip 0.03in}|@{\hskip 0.015in}c@{\hskip 0.015in}|@{\hskip 0.015in}c@{}}
        \toprule
        Patch size & $\delta1$ $\uparrow$ & $\delta2$ $\uparrow$ & $\delta3$ $\uparrow$ & RMSE (cm)$\downarrow$ & AbsRel (cm)$\downarrow$ \\
        \midrule
        $11 \times 11$ & 0.717 & \textbf{0.841} & \textbf{0.903} & 5.675 & 3.498 \\
        $21 \times 21$ & \textbf{0.720} & 0.840 & 0.895 & \textbf{5.281} & \textbf{3.295} \\
        $31 \times 31$ & 0.657 & 0.821 & 0.895 & 6.123 & 4.060 \\
        \bottomrule
    \end{tabular}
    \caption{Depth estimation accuracy for different patch sizes on the synthesized testing set. We report metrics commonly used in prior works~\cite{eigen2014depth,si2023fully,yang2022deep}. Detailed calculations of these metrics are in the supplementary.}
    \label{tab:exp:ablation}
\end{table}

\subsection{Results on Synthetic and Real Data}
\label{sec:experiment:synthetic}

\begin{table*}[htb]
    \centering
    \begin{tabular}{@{}l|l|c|c|ccc|c|c@{}}
        \toprule
        \multicolumn{2}{@{}l|}{Method} & Venue'Year & \# images & $\delta1$ $\uparrow$ & $\delta2$ $\uparrow$ & $\delta3$ $\uparrow$ & RMSE (cm) $\downarrow$ & AbsRel (cm) $\downarrow$ \\
        \midrule
        \multirow{3}{*}{\rotatebox[origin=c]{90}{Sparse}} & Focal Track~\cite{guo2017focal} & ICCV'2017 & 2 & 0.588 & 0.784 & 0.874 & 6.308 & 4.640 \\
        & Tang \etal~\cite{tang2017depth} & CVPR'2017 & 2 & 0.663 & 0.790 & 0.878 & 6.737 & 4.346 \\
        & \textbf{Ours} & - & 2 & \textbf{0.720} & \textbf{0.840} & \textbf{0.895} & \textbf{5.281} & \textbf{3.295} \\
        \midrule
        \multirow{6}{*}{\rotatebox[origin=c]{90}{Dense}} & PhaseCam3D~\cite{wu2019phasecam3d} & ICCP'2019 & 2 & 0.405 & 0.646 & 0.775 & 9.883 & 8.053 \\
        & DefocusNet~\cite{maximov2020focus} & CVPR'2020 & 5 & 0.657 & 0.847 & 0.908 & 6.092 & 4.548 \\
        & DFV-DFF~\cite{yang2022deep} & CVPR'2022 & 5 & 0.518 & 0.762 & 0.868 & 8.298 & 6.707 \\
        & DEReD~\cite{si2023fully} & CVPR'2023 & 5 & 0.536 & 0.778 & 0.874 & 7.779 & 5.977 \\
        & \textbf{Ours-W} & - & 2 & 0.628 & 0.812 & 0.885 & 6.297 & 4.525 \\
        & \textbf{Ours-PP} & - & 2 & \textbf{0.806} & \textbf{0.906} & \textbf{0.945} & \textbf{3.992} & \textbf{2.691} \\
        \bottomrule
    \end{tabular}
    \caption{Depth prediction accuracy on the synthetic testing set. The proposed algorithm has the best performance compared with the state-of-the-art algorithms on all metrics, with \emph{Ours} leading on sparse depth maps and \emph{Ours-PP} leading on dense depth maps. Details of the metrics are provided in the supplementary.}
    \label{tab:exp:syntheticresult}
\end{table*}

\begin{figure*}[htb]
    \centering
    \includegraphics[width=1.00\linewidth]{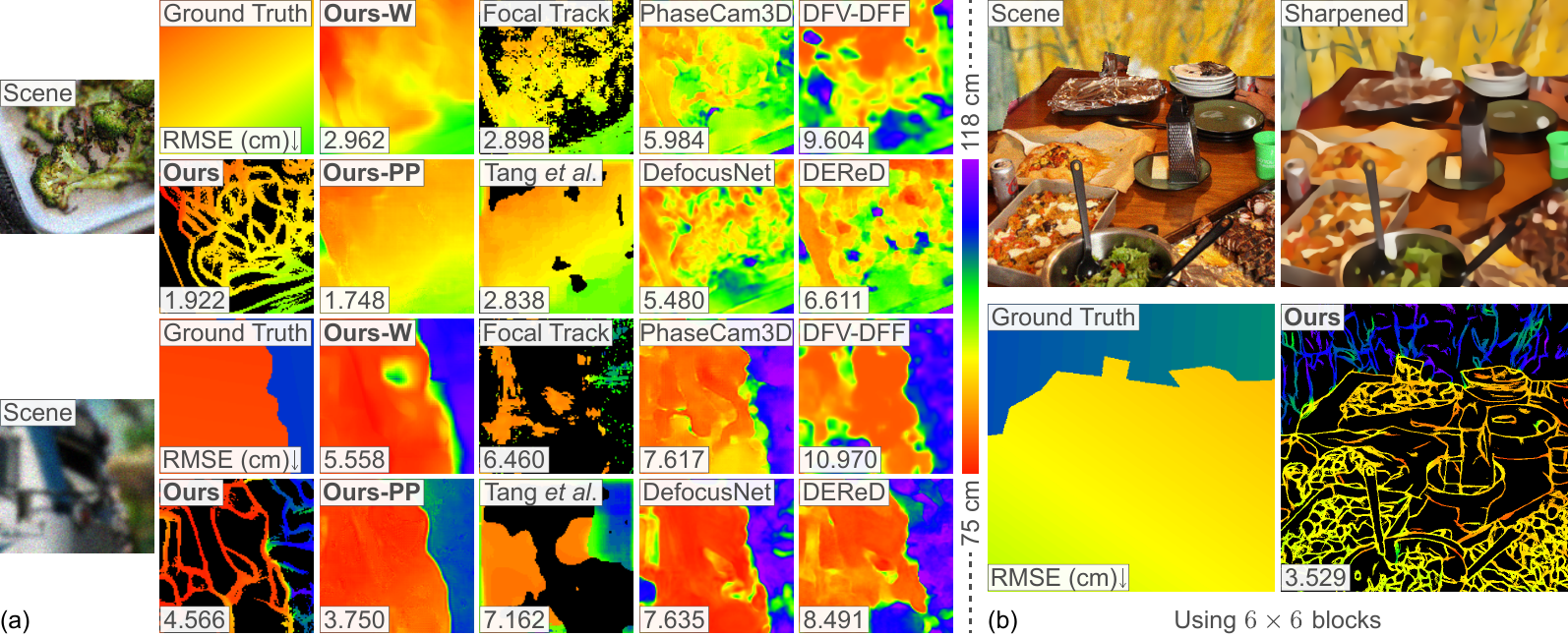}
    \caption{Depth maps from the synthetic testing set. (a) Our method can robustly predict sparse depth maps along boundaries for noisy input images. The sparse depth map can be effectively densified using two methods, \ie, \emph{Ours-W} and \emph{Ours-PP}. \emph{Ours-PP} achieves the highest visual quality and accuracy among all methods. A detailed explanation of the densification methods is in the supplementary. (b) Sample results of larger input images. Our method can handle images with higher resolution. It divides the input images into $147\times 147$ blocks and processes each block individually. A detailed description is in the supplementary.}
    \label{fig:exp:syntheticresult}

    \vspace{0.20in}
    
    \includegraphics[width=1.00\linewidth]{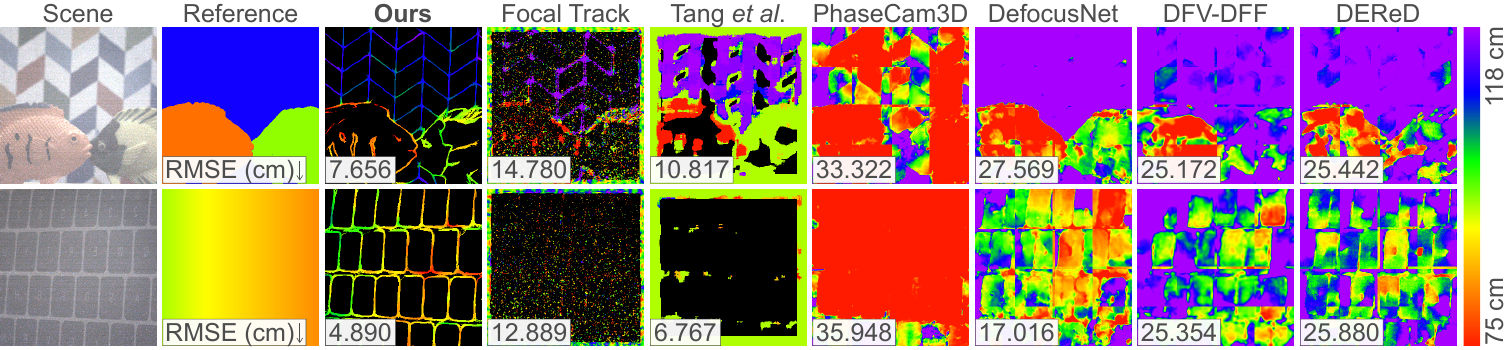}
    \caption{Depth maps on real-world images. Our method generates sparse depth maps of the highest visual quality and accuracy. The reference depth map is generated from manual measurements. The inset numbers are calculated according to the reference depth map.}
    \label{fig:exp:realresult}
\end{figure*}

We compare our method with various state-of-the-art DfD algorithms, including both analytical~\cite{guo2017focal,tang2017depth} and learning-based~\cite{wu2019phasecam3d, maximov2020focus, yang2022deep, si2023fully}. These algorithms are not originally designed for noisy images. We repurpose and retrain them using the same training data described in \cref{sec:experiment:dataset} to perform on noisy images. Our method can also output dense depth maps by assigning the depth values to wedges. Additionally, a post-processing can be adopted as a densifier. The notations \emph{Ours}, \emph{Ours-W}, and \emph{Ours-PP} refer to the sparse depth maps, dense depth maps from \methodname, and dense depth maps generated from the sparse depth maps using a U-Net~\cite{ronneberger2015u} as post-processing, respectively. More details about the densification of depth maps are in the supplementary.
The quantitative comparison on the testing set is shown in \cref{tab:exp:syntheticresult} with sample depth maps shown in \cref{fig:exp:syntheticresult}. 
Our model clearly achieves the best performance on all metrics and visually, with \emph{Ours} and \emph{Ours-PP} outperforming other methods on sparse and dense depth maps, respectively.
Besides images with standard $147\times 147$ resolution, our method can also handle larger images by dividing them into $147\times 147$ blocks to process individually. Additional details on how we merge the results of each block can be found in the supplementary. We show a sample result of input images with $587 \times 587$ resolution in \cref{fig:exp:syntheticresult}b.

We also build a prototype camera with a deformable lens similar to the one in Guo \etal~\cite{guo2017focal}, and use it to capture low-light, differently defocused image pairs or stacks to test the algorithms' performance on real-world data. \Cref{fig:exp:realresult} compares the sample depth maps from different methods. Depth maps from the proposed method demonstrate the highest visual quality. More details and results of the real-world experiments are in the supplementary.

\vspace{0.2in}
\noindent \textbf{Acknowledgement.} We thank Professors Emma Alexander, Stanley Chan, David Inouye, and Xiaoqian Wang for their valuable feedback. The work was partly supported by the U.S. National Science Foundation award CCF-2431505.

{
    \small
    \bibliographystyle{ieeenat_fullname}
    \bibliography{main}

\begin{thebibliography}{9}

    \bibitem{wang2022smish}
    Xueliang Wang, Honge Ren, and Achuan Wang. Smish: A novel activation function for deep learning methods. \textit{Electronics}, 11(4):540, 2022.

    \bibitem{wang2021translating}
    Zelun Wang and Jyh-Charn Liu. Translating math formula images to LaTeX sequences using deep neural networks with sequence-level training. \textit{International Journal on Document Analysis and Recognition (IJDAR)}, 24(1):63--75, 2021.
    
    \end{thebibliography}
}

\clearpage
\appendix
\renewcommand{\thepage}{S\arabic{page}}
\renewcommand{\thetable}{S\arabic{table}}
\renewcommand{\thefigure}{S\arabic{figure}}
\renewcommand{\theequation}{S\arabic{equation}}
\renewcommand{\thesection}{S\arabic{section}}
\setcounter{page}{1}
\setcounter{table}{0}
\setcounter{figure}{0}
\setcounter{equation}{0}
\setcounter{section}{0}
\maketitlesupplementary

\section{Calculation of illuminance}
\label{sec:illuminance}

To obtain the equivalent illuminance in \cref{tab:related:noise}, we first calculate the corresponding photon level $\alpha$ by solving:
\begin{equation}
    \frac{\sqrt{\alpha+\sigma^2}}{\alpha} \times 255 = \mathit{\mathrm{SD}}_{\text{LSB}}.
\end{equation}
Then compute the solid angle of a pixel $\Omega_{\text{pix}}$:
\begin{equation}
    \Omega_{\text{pix}} = \frac{A_{\text{pix}}}{f^2},
\end{equation}
where $A_{\text{pix}} = \left( 5.93 \times 10^{-6} \right)^2 \text{ m$^2$}$ is the area of one pixel. $P_{\text{pix}}$ is the power received by one pixel:
\begin{equation}
    P_{\text{pix}} = \frac{\alpha}{t \cdot \mathit{\mathrm{QE}}} \cdot \frac{hc}{\lambda_G},
\end{equation}
where $t = 1/200 \text{ s}$ is the exposure time, $\lambda_G = 532 \text{ nm}$ is the wavelength of green light, $\mathit{\mathrm{QE}} = 0.73$ is the quantum efficiency at $\lambda_G = 532 \text{ nm}$, $h = 6.626 \times 10^{-34} \text{ J$\cdot$s}$ is the Planck's constant, $c = 3 \times 10^8 \text{ m/s}$ is the speed of light. $A_{\text{aperture}}$ is the area of the aperture:
\begin{equation}
    A_{\text{aperture}} = \pi \Sigma^2.
\end{equation}
$f/\#$ is the f-number:
\begin{equation}
    f/\# = \frac{f}{2 \Sigma},
\end{equation}
and we choose $f/5.6$ in our calculation. $K_m = 683 \text{ lm/W}$ is the maximum possible value of photopic luminous efficacy of radiation. $V\left(\lambda_G\right) = 0.83$ is the photopic luminous efficiency function at $\lambda_G$. Finally, the illuminance is computed via:
\begin{equation}
    \begin{aligned}
        E_{\text{lux}} =& \frac{2 \pi}{\Omega_{\text{pix}}} \cdot \frac{P_{\text{pix}}}{A_{\text{aperture}}} \\
        =& \frac{8 \left(f/\#\right)^2}{A_{\text{pix}}} \cdot \frac{\alpha}{t \cdot \mathit{\mathrm{QE}}} \cdot \frac{hc}{\lambda_G} \cdot K_m \cdot V(\lambda_G)
    \end{aligned}.
\end{equation}

\section{Further discussion on the image model}
\label{sec:add_camera}

\subsection{Image rendering}
\label{sec:add_camera:defocus_tex}

The captured image $I(\boldsymbol{x})$ theoretically results from Gaussian convolutions of defocus (\cref{eq:met:cam}) and smooth textures smoothness (\cref{eq:met:texsmoothness}), applied independently to a piecewise step function $\bar{Q} \left( \boldsymbol{x} \right)$, requiring two convolutions:
\begin{equation}
    \begin{aligned}
        I(\boldsymbol{x}) =& \bar{Q}(\boldsymbol{x}) \ast k\left(\boldsymbol{x}, \xi \right) \ast k\left(\boldsymbol{x}, \sigma(z) \right) \\
        =& \bar{Q}(\boldsymbol{x}) \ast \left( k\left(\boldsymbol{x}, \xi \right) \ast k\left(\boldsymbol{x}, \sigma(z) \right) \right) \\
        =& \bar{Q}(\boldsymbol{x}) \ast k\left(\boldsymbol{x}, \sqrt{\sigma(z)^2 + \xi^2} \right)
    \end{aligned}.
\end{equation}
which leads to \cref{eq:met:imgrender}.

\subsection{Image formation model}
\label{sec:add_camera:thin_lens}

\begin{figure}
    \centering
    \includegraphics[width=1.00\linewidth]{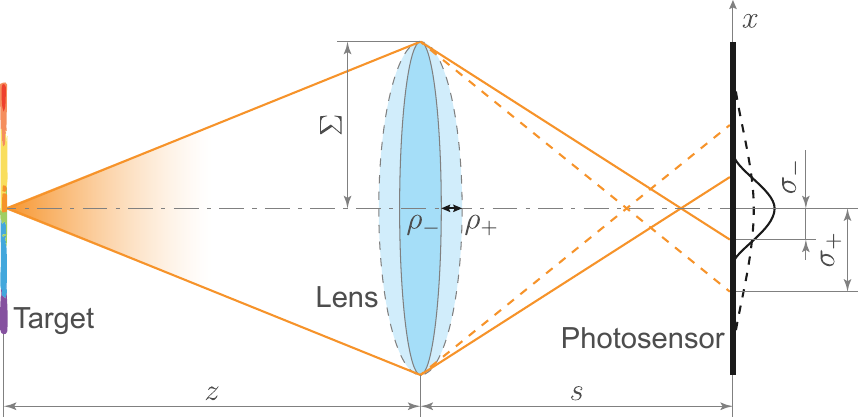}
    \caption{Image formation model. Consider a deformable lens that can change its optical power from $\rho_-$ to $\rho_+$. The point spread function of a fixed target changes its width from $\sigma_-$ to $\sigma_+$ according to the thin-lens law.}
    \label{fig:add_camera:thin_lens}
\end{figure}

To complement the mathematical derivation of our DfD equation and visualize key terms used in depth estimation, the thin lens model with a deformable lens is shown in \cref{fig:add_camera:thin_lens}.

\section{Derivation of distance maps}
\label{sec:dist_map}

\begin{figure}
    \centering
    \includegraphics[width=0.62\linewidth]{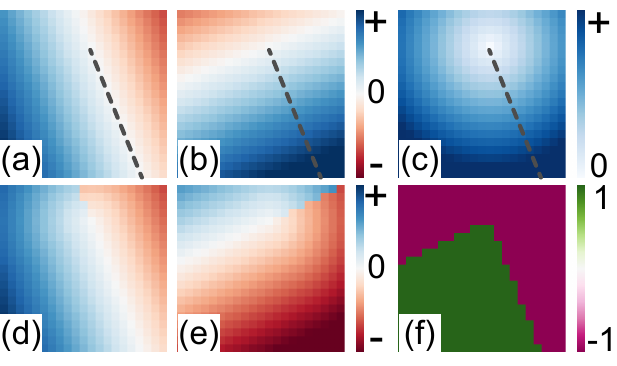}
    \caption{Additional visualizations from the sample \methodname representation in \cref{fig:met:dist}. (a-b) The signed distance maps in the radial and axial directions to the starting edge of the bottom wedge, $r_{11} \left( \boldsymbol{x}; \boldsymbol{\Psi} \right)$ and $a_{11} \left( \boldsymbol{x}; \boldsymbol{\Psi} \right)$, respectively. The location of the related boundary is noted as the dotted line. (c) The unsigned scaled distance map to the vertex $\boldsymbol{p}_1$, $v_{11} \left( \boldsymbol{x}; \boldsymbol{\Psi} \right)$, in this sample $w=1$. The location of the related boundary is noted as the dotted line. (d-e) The signed distance map to the starting and ending edges of the bottom wedge, $d_{11} \left( \boldsymbol{x}; \boldsymbol{\Psi} \right)$ and $d_{12} \left( \boldsymbol{x}; \boldsymbol{\Psi} \right)$, where $d_{11} \left( \boldsymbol{x}; \boldsymbol{\Psi} \right)$ is calculated through the map in (a-c). (f) The indicator function of the bottom wedge, $\chi_1 \left( \boldsymbol{x}; \boldsymbol{\Psi} \right)$.}
    \label{fig:dist_map:dist}
\end{figure}

We first introduce $d_{it} \left( \boldsymbol{x}; \boldsymbol{\Psi} \right)$, $t=1,2$, the signed distance map to the starting ($t=1$) or ending ($t=2$) edge of the $i$th wedge (\cref{fig:dist_map:dist}d-e):
\begin{equation}
    d_{it} \left( \boldsymbol{x}; \boldsymbol{\Psi} \right) \hspace{-0.03in} = \hspace{-0.04in}
    \begin{cases}
        r_{it} \left( \boldsymbol{x}; \boldsymbol{\Psi} \right) &\mkern-65mu \text{if $a_{it} \left( \boldsymbol{x}; \boldsymbol{\Psi} \right) \geq 0$}, \\
        \left[ 2 H \left( r_{it} \left( \boldsymbol{x}; \boldsymbol{\Psi} \right) \right) -1 \right] \cdot v_{it} \left( \boldsymbol{x}; \boldsymbol{\Psi} \right) &\mkern-12mu \text{otherwise},
    \end{cases}
\end{equation}
where $r_{it} \left( \boldsymbol{x}; \boldsymbol{\Psi} \right)$ is the signed distance map in the radial direction of the edge (\cref{fig:dist_map:dist}a):
\begin{equation}
    r_{it} \left( \boldsymbol{x}; \boldsymbol{\Psi} \right) \hspace{-0.006in} = \hspace{-0.006in} - \hspace{-0.005in} \left( x \hspace{-0.005in} - \hspace{-0.005in} x_i \right) \sin \left( \theta_{it} \right) \hspace{-0.004in} + \hspace{-0.004in} \left( y \hspace{-0.005in} - \hspace{-0.005in} y_i \right) \cos \left( \theta_{it} \right)\hspace{-0.005in},
\end{equation}
$a_{it} \left( \boldsymbol{x}; \boldsymbol{\Psi} \right)$ is the signed distance map in the axial direction of the edge (\cref{fig:dist_map:dist}b):
\begin{equation}
    a_{it} \left( \boldsymbol{x}; \boldsymbol{\Psi} \right) = \left( x - x_i \right) \cos \left( \theta_{it} \right) + \left( y - y_i \right) \sin \left( \theta_{it} \right),
\end{equation}
and $v_{it} \left( \boldsymbol{x}; \boldsymbol{\Psi} \right)$ is the unsigned scaled distance map to the vertex $\boldsymbol{p}_i$ (\cref{fig:dist_map:dist}c):
\begin{equation}
    v_{it} \left( \boldsymbol{x}; \boldsymbol{\Psi} \right) = \sqrt{\left( r_{it} \left( \boldsymbol{x}; \boldsymbol{\Psi} \right) \right)^2 + w^2 \cdot \left( a_{it} \left( \boldsymbol{x}; \boldsymbol{\Psi} \right) \right)^2},
    \label{eq:dist_map:dist2vertex}
\end{equation}
where $w$ is a scale factor. 

Then the signed distance map of the $i$th wedge $d_i \left( \boldsymbol{x}; \boldsymbol{\Psi} \right)$ is computed via:
\begin{equation}
    d_i \hspace{-0.022in} \left( \boldsymbol{x}; \hspace{-0.010in} \boldsymbol{\Psi} \right) \hspace{-0.028in} = \hspace{-0.028in} \min \hspace{-0.020in} \left( \left|d_{i1} \hspace{-0.023in} \left( \boldsymbol{x}; \hspace{-0.010in} \boldsymbol{\Psi} \right)\right|\hspace{-0.01in}, \hspace{-0.020in} \left|d_{i2} \hspace{-0.023in} \left( \boldsymbol{x}; \hspace{-0.010in} \boldsymbol{\Psi} \right)\right| \right) \hspace{-0.012in} \cdot \hspace{-0.012in} \chi_i \hspace{-0.026in} \left( \boldsymbol{x}; \hspace{-0.010in} \boldsymbol{\Psi} \right)\hspace{-0.017in},
    \label{eq:dist_map:signed_dist}
\end{equation}
where $\chi_i \left( \boldsymbol{x}; \boldsymbol{\Psi} \right)$ is a indicator function indicating whether pixel $\boldsymbol{x}$ is inside of the $i$th wedge:
\begin{equation}
    \chi_i \hspace{-0.01in} \left( \boldsymbol{x}; \boldsymbol{\Psi} \right) \hspace{-0.03in} = \hspace{-0.03in}
    \begin{cases}
        1 & \hspace{-0.051in} \text{if $\theta_{i1} \hspace{-0.011in} \leq \hspace{-0.011in} \arctan2 \left( \boldsymbol{x} \hspace{-0.012in} - \hspace{-0.012in} \boldsymbol{p}_i \right) \hspace{-0.011in} \leq \hspace{-0.011in} \theta_{i2}$}, \\
        -1 & \hspace{-0.051in} \text{otherwise}.
    \end{cases}
\end{equation}

Finally, the unsigned distance map $u \left( \boldsymbol{x}; \boldsymbol{\Psi} \right)$, mentioned in \cref{eq:met:bndrypatch}, is calculated by:
\begin{equation}
    \begin{aligned}
        u \left( \boldsymbol{x}; \boldsymbol{\Psi} \right) = \min & \left( \left|d_i \left( \boldsymbol{x}; \boldsymbol{\Psi} \right)\right|, \right. \\
        &\mkern5mu \left. \left|d_{i+1} \left( \boldsymbol{x}; \boldsymbol{\Psi} \right)\right|, \cdots, \left|d_l \left( \boldsymbol{x}; \boldsymbol{\Psi} \right)\right| \right), \\
        &\mkern-47mu \text{when $M_i \left( \boldsymbol{x} \right) = 1$},
    \end{aligned}
    \label{eq:dist_map:unsigned_dist}
\end{equation}
where $M_i \left( \boldsymbol{x} \right) = 1$ is the mask for the unoccluded $i$th wedge:
\begin{equation}
    M_i \hspace{-0.005in} \left( \boldsymbol{x} \right) \hspace{-0.01in} = \hspace{-0.01in} H \hspace{-0.005in} \left( d_i \hspace{-0.005in} \left( \boldsymbol{x}; \hspace{-0.005in} \boldsymbol{\Psi} \right) \right) \hspace{-0.001in} \prod_{j=i+1}^l \hspace{-0.001in} \left[ 1 \hspace{-0.005in} - \hspace{-0.005in} H \hspace{-0.005in} \left( d_j \hspace{-0.005in} \left( \boldsymbol{x}; \hspace{-0.005in} \boldsymbol{\Psi} \right) \right) \right].
    \label{eq:dist_map:unoccluded_mask}
\end{equation}

\section{Details of implementation}
\label{sec:net_arch}

\subsection{Local stage architecture}
\label{sec:net_arch:local}

\Cref{tab:net_arch:local} lists the hyperparameters of the convolutional neural network (CNN) of the local stage shown in \cref{fig:met:framework}. We adopt the Smish function as the activation function for each layer~\citeS{wang2022smish}:
\begin{equation}
    \mathrm{Smish}\left( x \right) = x \cdot \tanh \left[ \ln \left( 1 + \mathrm{sigmoid} \left( x \right)\right) \right],
\end{equation}
which we find to be more stable and accurate in our experiments. 

\begin{table}[htb]
    \centering
    \begin{tabular}{@{}lcc@{}}
        \toprule
        Layer & Specification & Output \\
        \midrule
        Conv2d & $7 \times 7, 64$ & $21 \times 21 \times 64$ \\
        MaxPool2d & $3 \times 3$ & $11 \times 11 \times 64$ \\
        ResBlock & $\begin{bmatrix}
            3 \times 3, 96 \\
            3 \times 3, 96
        \end{bmatrix}$ & $11 \times 11 \times 96$ \\
        MaxPool2d & $3 \times 3$ & $6 \times 6 \times 96$ \\
        ResBlock & $\begin{bmatrix}
            3 \times 3, 256 \\
            3 \times 3, 256
        \end{bmatrix}$ & $6 \times 6 \times 256$ \\
        ResBlock & $\begin{bmatrix}
            3 \times 3, 384 \\
            3 \times 3, 384
        \end{bmatrix}$ & $6 \times 6 \times 386$ \\
        ResBlock & $\begin{bmatrix}
            3 \times 3, 256 \\
            3 \times 3, 256
        \end{bmatrix}$ & $6 \times 6 \times 256$ \\
        MaxPool2d & $2 \times 2$ & $3 \times 3 \times 256$ \\
        Linear & - & $1024$ \\
        Linear & - & $10$ \\
        \bottomrule
    \end{tabular}
    \caption{CNN architecture of the local stage.}
    \label{tab:net_arch:local}
\end{table}

\subsection{Global stage architecture}
\label{sec:net_arch:global}

\begin{table}[htb]
    \centering
    \begin{tabular}{@{}lc@{}}
        \toprule
        Item & Value \\
        \midrule
        Dimension of each feature vector & 128 \\
        Number of sub-encoder-layers & 8 \\
        Number of heads in multi-head attention & 8 \\
        Dimension of the feedforward network model & 256 \\
        \bottomrule
    \end{tabular}
    \caption{Transformer Encoder details in the global stage. It takes in the \methodname representation parameters and does not have access to the input image pair in the inference.}
    \label{tab:net_arch:global}
\end{table}

\Cref{tab:net_arch:global} lists the architecture of the Transformer Encoder of the global stage. In this stage, each parameter pair, $\boldsymbol{\Psi}_{\pm}^{\boldsymbol{m}}$, is projected into a feature vector $\boldsymbol{v}^{\boldsymbol{m}} \in \mathbb{R}^v$ and added with a positional encoding vector $\boldsymbol{E}^{\boldsymbol{m}} = \begin{bmatrix}
    E^{\left(m,n,1\right)} & \cdots & E^{\left(m,n,v\right)}
\end{bmatrix}^{\top} \in \mathbb{R}^v$. The 2D positional encoding vector follows the design by Zhang and Liu~\citeS{wang2021translating}:
\begin{equation}
    E^{\left(m,n,q\right)} =
    \begin{cases}
        \sin \left( \frac{m}{10000^{\frac{2q}{v}}} \right) & \text{if $q$ is even and $q \leq \frac{v}{2}$}, \\
        \cos \left( \frac{m}{10000^{\frac{2q+2}{v}}} \right) & \text{if $q$ is odd and $q \leq \frac{v}{2}$}, \\
        \sin \left( \frac{n}{10000^{\frac{2q}{v}-1}} \right) & \text{if $q$ is even and $q > \frac{v}{2}$}, \\
        \cos \left( \frac{n}{10000^{\frac{2q+2}{v}-1}} \right) & \text{if $q$ is odd and $q > \frac{v}{2}$}.
    \end{cases}
\end{equation}

\subsection{Loss functions for local and global stages training}
\label{sec:net_arch:loss}

As shown in \cref{eq:met:local-obj,eq:met:global-obj}, we use comprehensive loss functions to train the CNN of the local stage and the Transformer Encoder of the global stage, respectively. For local stage training, three terms $l_{1-3}$ regularize the \methodname prediction as following:
\begin{equation}
    \begin{aligned}
        l_1 =& \left\|c \left( \boldsymbol{x}; \boldsymbol{\Psi}^{\boldsymbol{m}}_\pm \right) - P_{\text{clean},\pm}^{\boldsymbol{m}} \left( \boldsymbol{x} \right)\right\|^2 \mkern+59mu \text{(color error)},\\
        l_2 =& \left\|c^{\prime} \left( \boldsymbol{x}; \boldsymbol{\Psi}^{\boldsymbol{m}}_\pm \right) - P_{\text{clean},\pm}^{\boldsymbol{m}, \prime } \left( \boldsymbol{x} \right) \right\|^2 \mkern+7mu \text{(smoothness error)}, \\
        l_3 =& b \left( \boldsymbol{x}; \boldsymbol{\Psi}^{\boldsymbol{m}}_\pm, \delta \right) \cdot u^{\boldsymbol{m}} \left( \boldsymbol{x} \right) \mkern+30mu \text{(boundary localization)},
    \end{aligned}
\end{equation}
where the terms $P_{\text{clean},\pm}^{\boldsymbol{m}} \left( \boldsymbol{x} \right)$ and $P_{\text{clean},\pm}^{\boldsymbol{m}, \prime } \left( \boldsymbol{x} \right)$ indicates the noiseless image patch and its derivative map from Sobel filtering, and $u^{\boldsymbol{m}}$ represents the unsigned distance map to the nearest true boundaries in the patch. For global stage training, seven terms $g_{1-7}$ comprehensively penalize the \methodname prediction as detailed below:
\begin{equation}
    \begin{aligned}
        g_1 =& \left\|c \left( \boldsymbol{x}; \left\{ \boldsymbol{\Omega}^{\boldsymbol{m}}, \boldsymbol{\eta}^{\boldsymbol{m}} \right) \right\} - P_{\text{clean}}^{\boldsymbol{m}} \left( \boldsymbol{x} \right) \right\|^2 \mkern+48mu \text{(color error)}, \\
        g_2 =& \left\|c \left( \boldsymbol{x}; \left\{ \boldsymbol{\Omega}^{\boldsymbol{m}}, \boldsymbol{\eta}^{\boldsymbol{m}} \right\} \right) - C^{\boldsymbol{m}} \left( \boldsymbol{x} \right) \right\|^2 \mkern+9mu \text{(color consistency)}, \\
        g_3 =& \left\|b \left( \boldsymbol{x}; \boldsymbol{\Omega}^{\boldsymbol{m}}, \delta \right) - B^{\boldsymbol{m}} \left( \boldsymbol{x} \right)\right\|^2 \mkern+14mu \text{(boundary consistency)}, \\
        g_4 =& \left\|c^{\prime} \left( \boldsymbol{x}; \left\{ \boldsymbol{\Omega}^{\boldsymbol{m}}, \boldsymbol{\eta}^{\boldsymbol{m}} \right\} \right) - P_{\text{clean}}^{\boldsymbol{m}, \prime} \left( \boldsymbol{x} \right) \right\|^2 \mkern-6mu \text{(smoothness error)}, \\
        g_5 =& \left\|c^{\prime} \left( \boldsymbol{x}; \left\{ \boldsymbol{\Omega}^{\boldsymbol{m}}, \boldsymbol{\eta}^{\boldsymbol{m}} \right\} \right) - C^{\boldsymbol{m}, \prime} \left( \boldsymbol{x} \right) \right\|^2 \\
        &\mkern+202mu \text{(smoothness consistency)}, \\
        g_6 =& b \left( \boldsymbol{x}; \boldsymbol{\Omega}^{\boldsymbol{m}}, \delta \right) \cdot u^{\boldsymbol{m}} \left( \boldsymbol{x} \right) \mkern+55mu \text{(boundary localization)}, \\
        g_7 =& \left\|\sum_{i=1}^l H \left( b_i \left( \boldsymbol{x}; \boldsymbol{\Omega}^{\boldsymbol{m}}, \delta \right) - \tau \right) \cdot z_i^{\boldsymbol{m}} - Z^{\ast,\boldsymbol{m}} \left( \boldsymbol{x} \right) \right\|^2 \\
        &\mkern+295mu \text{(depth error)},
    \end{aligned}
\end{equation}
where $Z^{\ast,\boldsymbol{m}} \left( \boldsymbol{x} \right)$ denotes the ground truth depth map of the patch.

\subsection{Weight scheduling}
\label{sec:net_arch:train_param}

We observe that dynamically varying the weights of each term in the loss functions, $\beta_{1-3}$ and $\gamma_{1-7}$ (\cref{eq:met:local-obj} and \cref{eq:met:global-obj} in the main paper), benefit the convergence of both the local and global stages.
In addition to the parameters listed in \cref{sec:experiment:config}, we list the full details of the training parameters here. 

\begin{figure}[htb]
\centering
\includegraphics[width=1.00\linewidth]{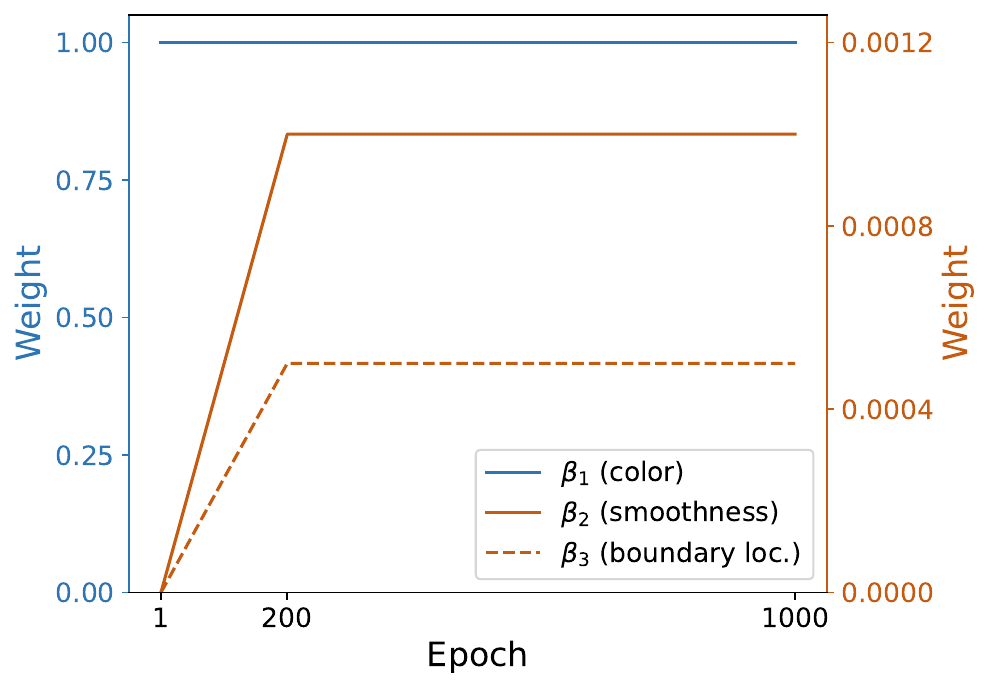}
\caption{Weight scheduling in the local stage training. There is one dynamic phase in the beginning. $\beta_2$ (for smoothness error) and $\beta_3$ (for boundary localization error) gradually increase to the final values in the first 200 epochs.}
\label{fig:net_arch:train_param:local}
\end{figure}

\begin{figure}[htb]
\centering
\includegraphics[width=1.00\linewidth]{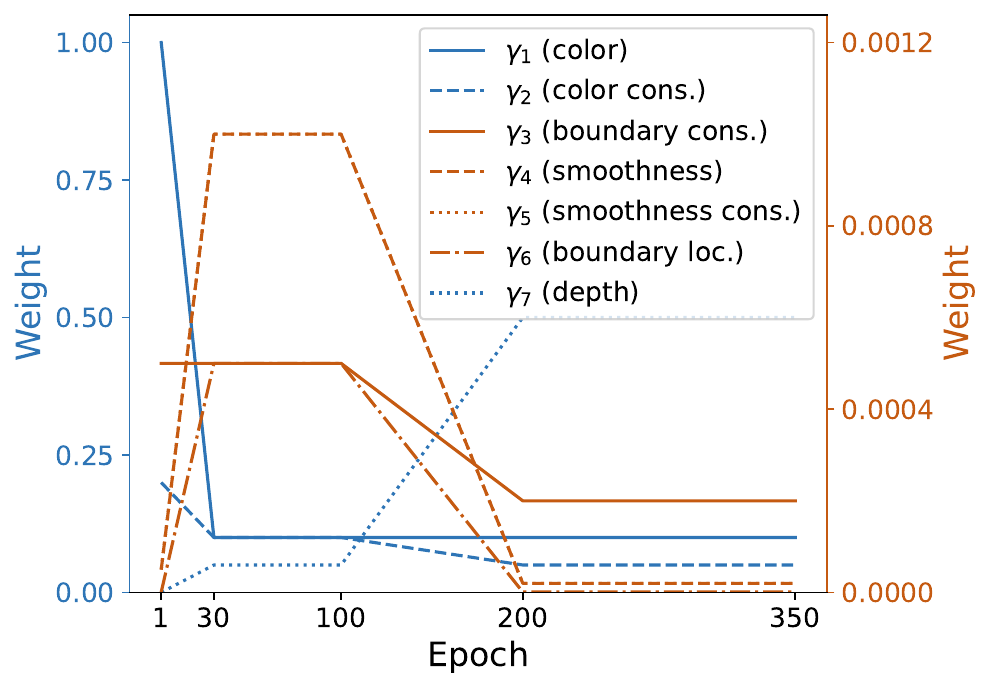}
\caption{Weight scheduling in the global stage training. There are two dynamic phases, 1--30 and 100--200 epochs individually. In the beginning, $\gamma_1$ (for color error) and $\gamma_2$ (for color consistency error) dominate the loss function, and $\gamma_3$ (for boundary consistency error) regularizes the consistency more at this phase. Then $\gamma_4$ (for smoothness error), $\gamma_5$ (for smoothness consistency error, overlapped with $\gamma_4$), and $\gamma_6$ (for boundary localization error) increase to refine the global color and boundary map. Finally, $\gamma_7$ (for depth error) leads the loss function to penalize the depth estimation.}
\label{fig:net_arch:train_param:global}
\end{figure}

We use $\lambda = 5 \times 10^{-3}$ in \cref{eq:met:ridge} and $w=1$ in \cref{eq:dist_map:dist2vertex} for both local and global stages. We apply dynamic loss function weights in \cref{eq:met:local-obj} and \cref{eq:met:global-obj} for local and global stages respectively. This strategy improves the stability and accuracy of the training. For each epoch, the weights are updated using linear interpolation, and the weight values are shown in \cref{fig:net_arch:train_param:local} and \cref{fig:net_arch:train_param:global}.

\subsection{Processing large images using blocks}
\label{sec:net_arch:block}

\begin{figure}[htb]
\centering
\includegraphics[width=0.55\linewidth]{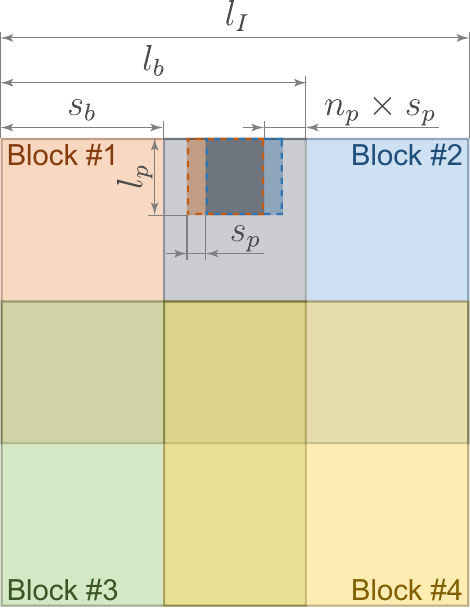}
\caption{Performing on large image using blocks. In this example, there are $2 \times 2$ blocks to cover the image. The final upper-right patch of \colorbox[HTML]{FBE5D6}{orange block} and the final upper-left patch of the \colorbox[HTML]{DEEBF7}{blue block} are indicated by the \colorbox[HTML]{F4B183}{dark orange square} and \colorbox[HTML]{9DC3E6}{dark blue square}, respectively}
\label{fig:net_arch:block}
\end{figure}

For large images, we divide them into blocks, as shown in \cref{fig:net_arch:block}. After estimating the depth for the patches within each block, the model combines them to generate the final depth maps. The number of blocks $n_b$ is calculated via:
\begin{equation}
    n_b = \left\lceil \frac{l_I - l_b}{s_b} + 1\right\rceil,
\end{equation}
where $l_I$ is the image side length, $l_b$ is the block side length, and $s_b$ is the block stride that is obtained through:
\begin{equation}
    s_b = l_b - l_p + \left( 1 - 2 n_p \right) s_p,
\end{equation}
where $l_p$ is the patch side length, $n_p$ is the number of marginal patches removed along the side length dimension, and $s_p$ is the patch stride. When $\left(2 n_p + 1\right)s_p >l_p$, the overlapped areas of blocks ensure that all patches are optimized with respect to all neighboring patches, mitigating the discontinuities between blocks.

\section{Improvement on DEReD training}
\label{sec:dered}

We observe that DEReD~\cite{si2023fully} struggles to be trained with the original loss function due to the lack of textures in our basic shape training set. To address this, we add two regularization terms: one for the $\ell^2$ norm of the depth map and the other for the $\ell^2$ norm of the depth map's first derivative. 

\section{Additional results}
\label{sec:add_result}

\subsection{Depth estimation metrics}
\label{sec:add_result:metric}

Here, we provide the computation of each metric used to quantify the depth estimation accuracy in the main paper. 

Given an estimated depth map $Z \left( \boldsymbol{x} \right)$ and the ground truth depth map $Z^{\ast} \left( \boldsymbol{x} \right)$, the RMSE and AbsRel metrics are calculated via:
\begin{equation}
    \text{RMSE} = \sqrt{\frac{\sum_{\boldsymbol{x}} \left\| Z \left( \boldsymbol{x} \right) - Z^{\ast} \left( \boldsymbol{x} \right) \right\|^2}{\left| \hat{Z} \left( \boldsymbol{x} \right) \right| }}
\end{equation}
and
\begin{equation}
    \text{AbsRel} = \frac{\sum_{\boldsymbol{x}} \frac{\left\| Z \left( \boldsymbol{x} \right) - Z^{\ast} \left( \boldsymbol{x} \right)\right\|}{Z^{\ast} \left( \boldsymbol{x} \right)}}{\left| \hat{Z} \left( \boldsymbol{x} \right) \right| },
\end{equation}
where $\left| \cdot \right|$ is the cardinality operator returning the number of pixels. For the $\delta$-threshold, we use the normalized maps of $Z \left( \boldsymbol{x} \right)$ and $Z^{\ast} \left( \boldsymbol{x} \right)$ with the range $[0,1]$, $\hat{Z} \left( \boldsymbol{x} \right)$ and $\hat{Z}^{\ast} \left( \boldsymbol{x} \right)$ respectively:
\begin{equation}
    \begin{aligned}
        & \hat{Z} \left( \boldsymbol{x} \right) = \frac{Z \left( \boldsymbol{x} \right) - Z_{\text{min}}}{Z_{\text{max}} - Z_{\text{min}}}, \\
        & \hat{Z}^{\ast} \left( \boldsymbol{x} \right) = \frac{Z^{\ast} \left( \boldsymbol{x} \right) - Z_{\text{min}}}{Z_{\text{max}} - Z_{\text{min}}}, \\
        & \delta i = \frac{\left| \max \left( \frac{\hat{Z} \left( \boldsymbol{x} \right)}{\hat{Z}^{\ast} \left( \boldsymbol{x} \right)}, \frac{\hat{Z}^{\ast} \left( \boldsymbol{x} \right)}{ \hat{Z} \left( \boldsymbol{x} \right)} \right) < \tau_n^i \right|}{\left| \hat{Z} \left( \boldsymbol{x} \right) \right|}, i=1,2,3,
    \end{aligned}
\end{equation}
where $Z_{\text{min}}$ and $Z_{\text{max}}$ are the minimum and maximum of the working range, and $\tau_n = 1.25$ is the threshold. 

\subsection{Ablation study on weight scheduling}
\label{sec:add_result:ablation}

We conducted an ablation study on different variations of weights in the loss function for the training of the global stage. Based on the configuration in \cref{fig:net_arch:train_param:global}, we mainly vary the scheduling of $\gamma_2$ (corresponding to color consistency error) and $\gamma_7$ (corresponding to depth error) to explore whether increasing the weights of the depth error can improve the performance, as shown in \cref{tab:add_result:config}. The results are in \cref{tab:add_result:ablation}. As our task requires accurate depth prediction at the correct boundary positions, a balanced weight between the depth error and the boundary error (Config 1) leads to the highest performance. 

\begin{table}[htb]
    \centering
    \begin{tabular}{@{}l|cc|c>{\columncolor{blue!20}[0pt]}c|>{\columncolor{orange!25}[4pt]}c>{\columncolor{blue!20}[0pt]}c@{}}
        \toprule
        \multirow{2}{*}{\shortstack[l]{Key\\epoch}} & \multicolumn{2}{c|}{Config 1 (base)} & \multicolumn{2}{c|}{Config 2} & \multicolumn{2}{c}{Config 3} \\
        & $\gamma_2$ & $\gamma_7$ & $\gamma_2$ & $\gamma_7$ & $\gamma_2$ & $\gamma_7$ \\
        \midrule
        1   & 0.2  & 0.0001 & 0.2  & 0.0001 & 0.1  & 0.0001 \\
        30  & 0.1  & 0.05   & 0.1  & 0.1    & 0.05 & 0.1    \\
        100 & 0.05 & 0.5    & 0.05 & 0.8    & 0.02 & 0.8    \\
        350 & 0.05 & 0.5    & 0.05 & 0.8    & 0.02 & 0.8    \\
        \bottomrule
    \end{tabular}
    \caption{Configurations of different weights for the ablation study. \emph{Config 1} is the base configuration in \cref{fig:net_arch:train_param:global}, and \emph{Config 2} increases the weight for the depth error, while \emph{Config 3} additionally decreases the weight for the color consistency error.}
    \label{tab:add_result:config}
\end{table}

\begin{table}[htb]
    \centering
    \begin{tabular}{@{}l@{\hskip 0.05in}|@{\hskip 0.05in}c@{\hskip 0.07in}c@{\hskip 0.07in}c@{\hskip 0.05in}|@{\hskip 0.035in}c@{\hskip 0.035in}|@{\hskip 0.035in}c@{}}
        \toprule
        Config & $\delta1$ $\uparrow$ & $\delta2$ $\uparrow$ & $\delta3$ $\uparrow$ & RMSE (cm)$\downarrow$ & AbsRel (cm)$\downarrow$ \\
        \midrule
        1 & \textbf{0.720} & \textbf{0.840} & \textbf{0.895} & \textbf{5.281} & \textbf{3.295} \\
        2 & 0.680 & 0.820 & 0.882 & 5.702 & 3.657 \\
        3 & 0.671 & 0.823 & 0.884 & 5.768 & 3.797 \\
        \bottomrule
    \end{tabular}
    \caption{Depth estimation accuracy for different configurations on the synthetic testing set. \emph{Config 1} leads to the highest accuracy on all metrics.}
    \label{tab:add_result:ablation}
\end{table}

\subsection{Evaluation with sparse masks}
\label{sec:add_result:more_eva}

\begin{table*}[htb]
    \centering
    \begin{tabular}{@{}l|l|c|c|ccc|c|c@{}}
        \toprule
        \multicolumn{2}{@{}l|}{Method} & Venue'Year & \# images & $\delta1$ $\uparrow$ & $\delta2$ $\uparrow$ & $\delta3$ $\uparrow$ & RMSE (cm) $\downarrow$ & AbsRel (cm) $\downarrow$ \\
        \midrule
        \multirow{4}{*}{\rotatebox[origin=c]{90}{Sparse}} & PhaseCam3D~\cite{wu2019phasecam3d} & ICCP'2019 & 2 & 0.408 & 0.669 & 0.805 & 9.115 & 7.715 \\
        & DefocusNet~\cite{maximov2020focus} & CVPR'2020 & 5 & 0.633 & 0.821 & 0.894 & 6.132 & 4.707 \\
        & DFV-DFF~\cite{yang2022deep} & CVPR'2022 & 5 & 0.486 & 0.747 & 0.862 & 8.312 & 6.815 \\
        & DEReD~\cite{si2023fully} & CVPR'2023 & 5 & 0.508 & 0.754 & 0.859 & 7.799 & 6.214 \\
        \bottomrule
    \end{tabular}
    \caption{Depth prediction accuracy of competing learn-based algorithms on the synthetic testing set after applying the same mask as ours for sparse depth map evaluation. According to \cref{tab:exp:syntheticresult}, our method achieves the highest performance.}
    \label{tab:add_result:syntheticresultmask}
    
    \vspace{0.15in}
    
    \begin{tabular}{@{}l|l@{\hskip 0.25in}|c|c|ccc|c|c@{}}
        \toprule
        \multicolumn{2}{@{}l|}{Method} & Venue'Year & \# images & $\delta1$ $\uparrow$ & $\delta2$ $\uparrow$ & $\delta3$ $\uparrow$ & RMSE (cm) $\downarrow$ & AbsRel (cm) $\downarrow$ \\
        \midrule
        \multirow{3}{*}{\rotatebox[origin=c]{90}{Sparse}} & Focal Track~\cite{guo2017focal} & ICCV'2017 & 2 & 0.503 & 0.709 & 0.857 & 7.794 & 6.103 \\
        & Tang \etal~\cite{tang2017depth} & CVPR'2017 & 2 & \textbf{0.726} & 0.832 & \textbf{0.902} & 5.991 & 3.537 \\
        & \textbf{Ours} & - & 2 & 0.714 & \textbf{0.836} & 0.886 & \textbf{5.007} & \textbf{3.255} \\
        \bottomrule
    \end{tabular}
    \caption{Depth prediction accuracy on synthetic large image testing set. Only competing methods capable of directly handling large images are shown. Our model adopts $3 \times 3$ blocks for this test.}
    \label{tab:add_result:syntheticresultlarge}
\end{table*}

In the main paper, we evaluate PhaseCam3D~\cite{wu2019phasecam3d}, DefocusNet~\cite{maximov2020focus}, DFV-DFF~\cite{yang2022deep}, and DEReD~\cite{si2023fully} with dense depth maps. For fairness, we re-evaluate these algorithms by masking out the same pixels as ours. See numbers in \cref{tab:add_result:syntheticresultmask}. Compared with the results of \emph{Ours} in \cref{tab:exp:syntheticresult}, our method universally performs the best.

\subsection{Additional results on synthetic images}
\label{sec:add_result:synthetic}

\begin{figure*}[htb]
\centering
\includegraphics[width=0.64\linewidth]{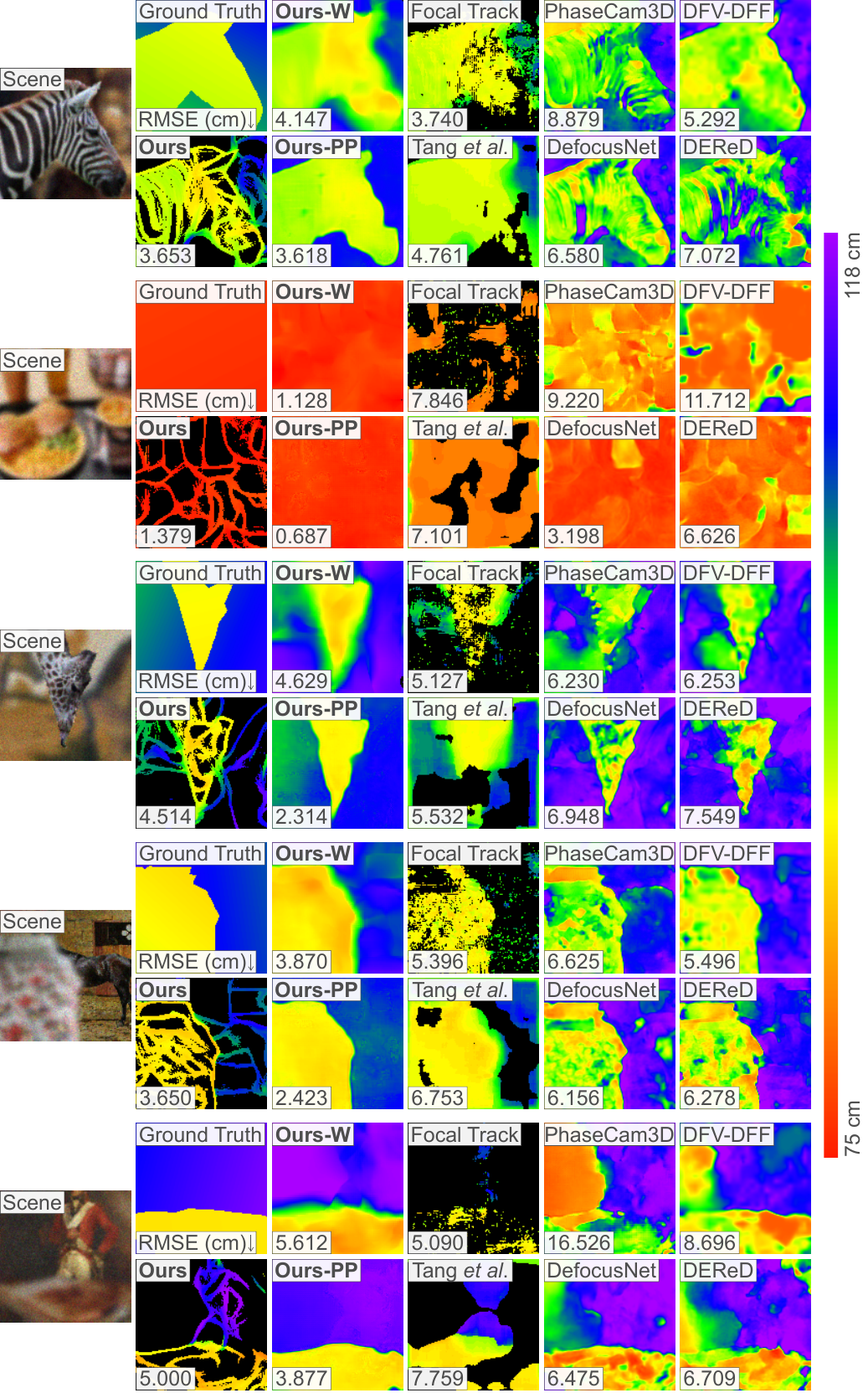}
\vspace{-0.04in}
\caption{Depth estimation on the synthetic testing set.}
\label{fig:add_result:syntheticresult}
\end{figure*}

\Cref{fig:add_result:syntheticresult} shows additional results on synthetic testing set. Compared to other algorithms, our method predicts the depth more accurately, especially when the images have abundant textures. 

\subsection{Additional results on synthetic large images}
\label{sec:add_result:big}

\begin{figure*}[htb]
\centering
\includegraphics[width=0.99\linewidth]{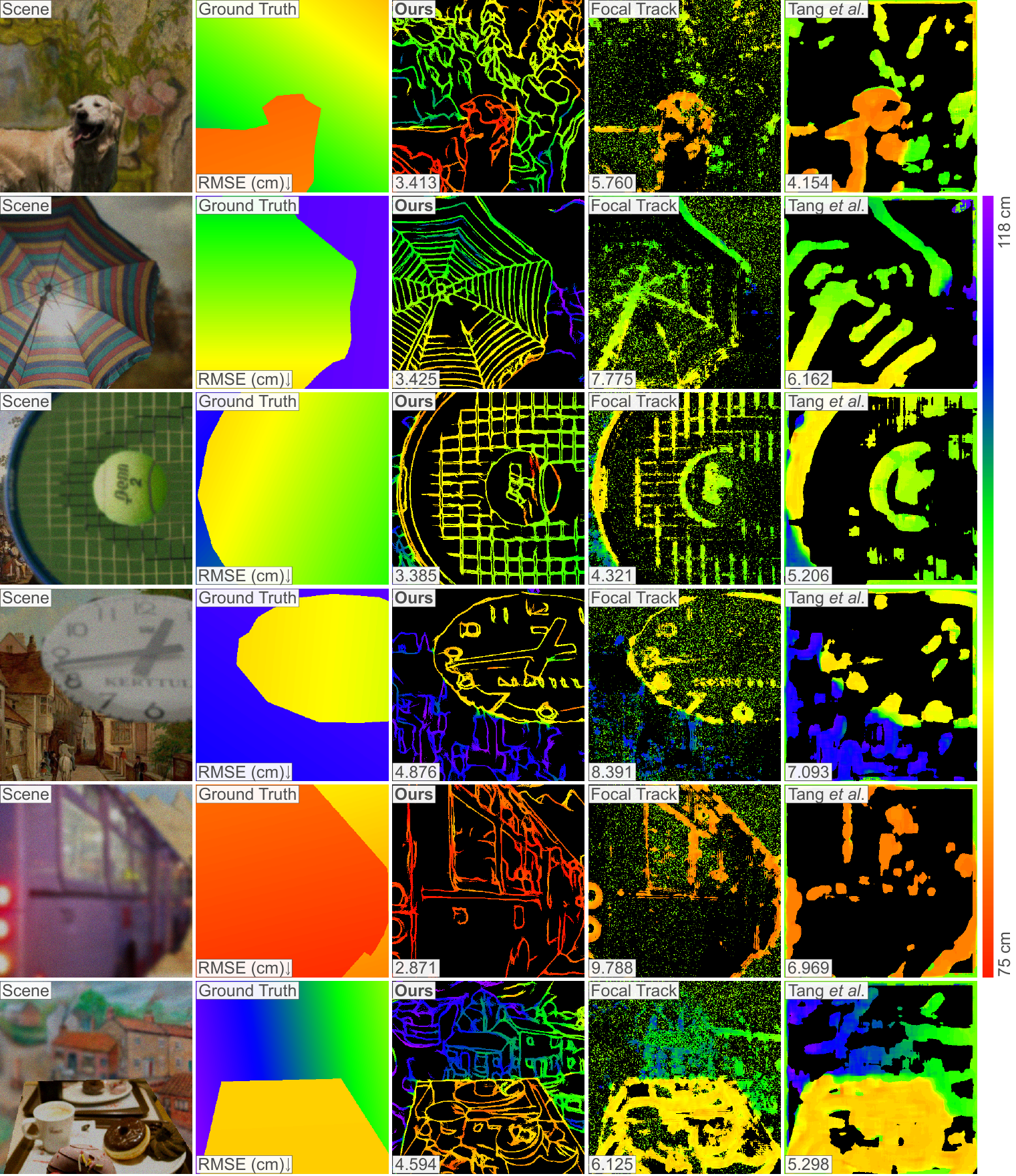}
\vspace{-0.11in}
\caption{Depth estimation on synthetic large images. Our method performs the best visual quality.}
\label{fig:add_result:big}
\vspace{-0.13in}
\end{figure*}

We evaluate our model on synthetic large images using $3 \times 3$ blocks. The quantitative comparison is shown in \Cref{tab:add_result:syntheticresultlarge}, where Focal Track~\cite{guo2017focal} and Tang \etal~\cite{tang2017depth} can process large images directly. \Cref{fig:add_result:big} presents a qualitative comparison on synthetic large images. Compared to other algorithms, our method predicts the depth more accurately.

\subsection{Additional results on real-world images}
\label{sec:add_result:real}

\begin{figure*}[htb]
\centering
\includegraphics[width=0.99\linewidth]{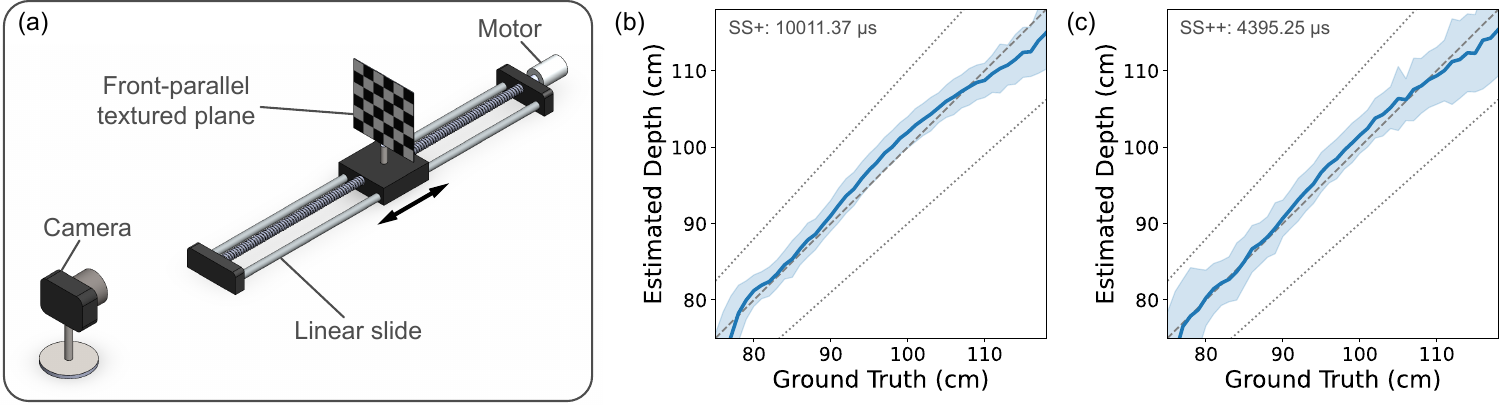}
\vspace{-0.05in}
\caption{Calibration of the real system. (a) Calibration setup. We place front-parallel textured planes accurately at known distances using a linear slide to collect input images with ground truth depth values. (b-c) Depth prediction accuracy after calibration at shutter speeds, \emph{SS+} and \emph{SS++}. The blue curve shows the mean predicted depth for each true depth. The vertical width of the color band indicates the standard deviation of the depth prediction error, which remains within 10$\%$ of the true depth (dotted line) across the working range.}
\label{fig:add_result:real:setup}
\end{figure*}

\begin{figure*}[htb]
\centering
\includegraphics[width=1.00\linewidth]{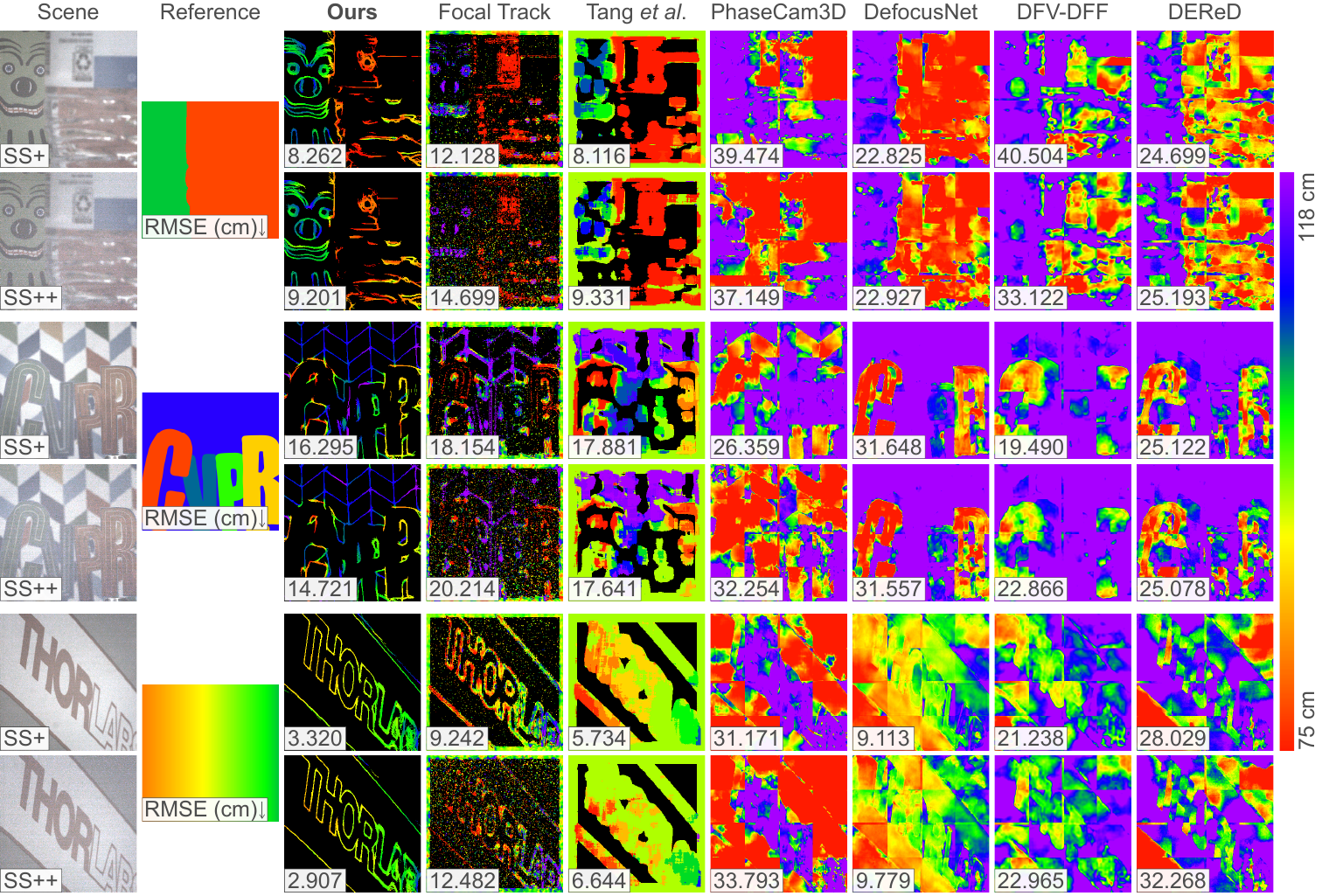}
\vspace{-0.20in}
\caption{Depth predictions from real-world images. Our method outperforms others in nearly all scenes. While Tang \etal~\cite{tang2017depth} slightly surpasses ours in the first scene at \emph{SS+}, our method performs better at \emph{SS++} and maintains the image structure more effectively, proving the robustness to high noise.}
\label{fig:add_result:real:comparison}
\end{figure*}

We build a prototype camera system that is similar to the setup in Guo \etal~\cite{guo2017focal}, including an Optotune EL-16-40-TC-VIS-5D-C to change the optical power and a FLIR Grasshopper 3 GS3-U3-23S6C-C camera. 

As the optical parameters of the physical system are different from the ones set for the synthetic data, we perform a linear mapping to calibrate depth estimation:
\begin{align}
    Z_{\text{output}} = \omega_0 + \omega_1 Z,
\end{align}
where $Z$ is the predicted depth value from our model, and $Z_{\text{output}}$ is the calibrated depth value that should match the actual object depths. We determine the parameters $\omega_0$ and $\omega_1$ using the following approach. We use a linear slide mounted with a front-parallel texture pad, as shown in \cref{fig:add_result:real:setup}. By moving the texture pad to different true distances $Z^{\ast}$, we obtain a series of mapping from the mean predicted depth value from our model $\{\Bar{Z}_i\}$ to the corresponding true depth $\{Z^{\ast}_i\}$. We solve the following linear regression problem:
\begin{equation}
    \begin{bmatrix}
        \vdots \\
        Z^{\ast}_i \\
        \vdots
    \end{bmatrix} = \omega_0 + \omega_1 \begin{bmatrix}
        \vdots \\
        \Bar{Z}_i \\
        \vdots
    \end{bmatrix}.
\end{equation}
The same mapping is applied to all other methods. 

We collect real data in an indoor environment with normal illumination, using two shutter speeds, 10011.37 \textmu s and 4395.25 \textmu s, denoted by \emph{SS+} and \emph{SS++}. The results in \cref{fig:exp:realresult} are with the shutter speed \emph{SS++}. Additional depth estimation results on real captured data are shown in \cref{fig:add_result:real:comparison}. Our method achieves the best performance in almost all scenes. Although Tang \etal~\cite{tang2017depth} performs slightly better than ours in the first scene at \emph{SS+}, ours outperforms it at \emph{SS++} and better retains the image structure. 

\section{Densification of depth maps}
\label{sec:dense}

\subsection{Dense depth maps from \methodname}
\label{sec:dense:blurryedges}

The proposed method aims to generate sparse depth maps along boundaries. However, we also experiment with generating dense depth maps by utilizing the \methodname representation. We compute a dense depth map using the following equation:
\begin{equation}
    Z \left( \boldsymbol{x} \right) =  \frac{\sum_{P^{\boldsymbol{m}}_{\pm} \ni \boldsymbol{x}} \sum_{i=1}^l
    M_i \left( \boldsymbol{x} \right) \cdot z_i^{\boldsymbol{m}}}{\sum_{P^{\boldsymbol{m}}_{\pm} \ni \boldsymbol{x}} \sum_{i=1}^l
    M_i \left( \boldsymbol{x} \right)}.
\end{equation}

We retrain the same global stage architecture for dense depth maps generation with an additional loss term $g_8$ in the objective function \cref{eq:met:global-obj} with the weight $\gamma_8$:
\begin{equation}
    g_8 = \left\|\sum_{i=1}^l
    M_i \left( \boldsymbol{x} \right) \cdot z_i^{\boldsymbol{m}} - Z^{\ast,\boldsymbol{m}} \left( \boldsymbol{x} \right) \right\|^2.
\end{equation}
The loss function penalizes the prediction error of the dense depth map compared to the ground truth. According to \cref{fig:add_result:syntheticresult}, the visual quality of the dense depth map depends on the distribution of boundaries in the images; the denser the boundaries are, the higher the quality of the depth maps. 

\subsection{Dense depth maps through post-processing}
\label{sec:dense:postproc}

We use a U-Net~\cite{ronneberger2015u} as a densifier that takes in the sparse depth maps and outputs the dense depth maps. For the training, the loss function optimizes the parameters through $\ell^2$ norm on the depth map and $\ell^2$ norm on its first derivative:
\begin{equation}
    \begin{aligned}
    \mathcal{L}_{\text{densify}} =& \nu_1 \mathbb{E}_{Z} \left( \left\| Z \left( \boldsymbol{x} \right) - Z^{\ast} \left( \boldsymbol{x} \right) \right\|^2 \right) \\
    &\mkern-4mu + \nu_2 \mathbb{E}_{Z} \left( \left\| Z^{\prime} \left( \boldsymbol{x} \right) - Z^{\ast, \prime} \left( \boldsymbol{x} \right)\right\|^2 \right),
    \end{aligned}
    \label{eq:dense:loss}
\end{equation}
where $\mathbb{E}_{Z}$ denotes the expectation over all dense depth maps, and $\nu_i$'s are the weight coefficients. \Cref{fig:add_result:syntheticresult} shows that the sparse depth map by our method is a sufficient output, and can be easily densified. 

{
    \small
    \makeatletter
    \renewcommand{\@biblabel}[1]{[S#1]}
    \makeatother
    
}

\end{document}